# Systems of natural-language-facilitated human-robot cooperation: A review

Rui Liu, Xiaoli Zhang*

*Department of Mechanical Engineering, Colorado School of Mines, Golden, CO 80401, USA*
*{rliu, xlzhang}@mines.edu*

**Abstract:** Natural-language-facilitated human-robot cooperation (NLC), in which natural language (NL) is used to share knowledge between a human and a robot for conducting intuitive human-robot cooperation (HRC), is continuously developing in the recent decade. Currently, NLC is used in several robotic domains such as manufacturing, daily assistance and health caregiving. It is necessary to summarize current NLC-based robotic systems and discuss the future developing trends, providing helpful information for future NLC research. In this review, we first analyzed the driving forces behind the NLC research. Regarding to a robot's cognition level during the cooperation, the NLC implementations then were categorized into four types "NL-based control, NL-based robot training, NL-based task execution, NL-based social companion" for comparison and discussion. Last based on our perspective and comprehensive paper review, the future research trends were discussed.

## 1. Introduction

### 1.1 Advantages

Natural-language-facilitated human-robot cooperation (NLC), in which a human uses either spoken or written instructions to communicate with a robot for task cooperation [1][2][3][4], has received increasing attention in human-involved robotics research. By using natural language (NL), human intelligence at high-level mental planning and robot capability at low-level physical execution are combined to perform an intuitive cooperation [5][6].

Compared with human-robot cooperation (HRC) using tactile indications such as contact location [7] and force strength [8][9][10], and visual indications such as body pose [11][12][13] and motion [14][15][16], HRC using NL indications has several advantages. First, NL makes the HRC *natural*. For the traditional methods mentioned above, the human involved in HRC needs to be trained to use certain actions/poses for making themselves understandable [17][18][19][20][21] . While in NLC, even non-expert users without prior training can cooperate with a robot by using human-like communication [3][22][23] . Second, the cooperation request is described *accurately*. The traditional methods using actions/poses only provide limited patterns to roughly describe cooperation requests due to the information loss in the action/pose simplification (such as using markers to simplify the actions) [24][25][26][27]. While in NLC, cooperation requests related to action, speed, tool and location are already defined in NL expressions [5][28][29][30]. With these expressions, cooperation requests for various task executions are described accurately. Third, NL transfers cooperation requests *efficiently*. The information-transferring method using actions/poses requires the design of informative patterns for different cooperation requests [24][25][26][27]. While existing languages, such as English, Chinese and German, already have standard linguistic structures, which contain abundant informative expressions to serve as patterns [31][32]. NL-based methods do not need to design specific informative patterns for various cooperation requests, making HRC efficient. Lastly, since the instructions are delivered orally instead of being physically involved, human hands are set free to perform more important executions [33]. Attracted by the above advantages, NLC has been widely explored in areas, including daily assistances [1][33][35], medical caregiving [36][37][38][39], manufacturing [5][40][41], indoor/outdoor navigation [2][42][43], social accompany [44][45][46] etc. Typical areas using NLC systems are shown in Fig. 1.

From a realization perspective, the pushing forces for recent NLC developments are concluded as natural language processing (NLP) developments and HRC developments.

### 1.2. Pushing force one: development of NLP

Recently, supported by machine learning technics in classification [47], clustering [48] and feature extraction [49], NLP has been developed from simply syntax-driven processing, which builds syntax representations of sentence

---

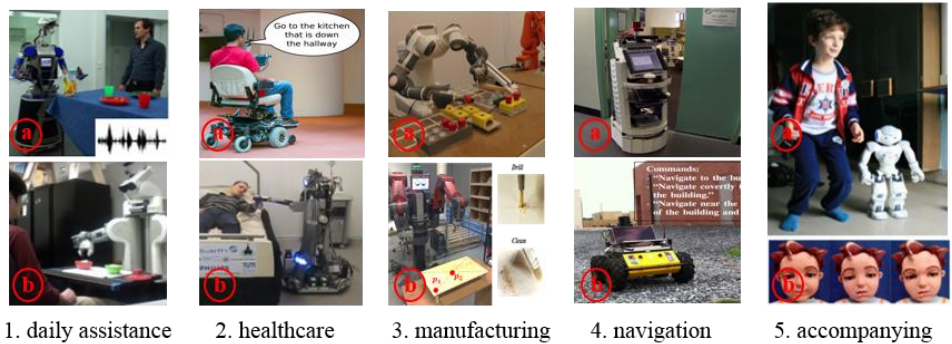

Fig. 1. Typical areas using NLC systems. 1.a[66] and 1.b[1] are NL-based robot daily assistance. 2.a [36] and 2.b [37] are NL-based healthcare. 3.a [41] and 3.b [5] are NL-based intelligence manufacturing. 4.a [42] and 4.b [43] are NL-based indoor/outdoor navigation. 5.a [67] and 5.b [45] are NL-based accompanying.

structures, to semantically-driven processing, which builds semantic networks for sentence meanings [50].

At the early stage (1950-2000) of NLP [51], a word-based understanding method was developed to enable a naïve word-symbol understanding by identifying single/multiple keywords [52][53], lexical affinities [54][55], and word/affinity occurrences [56][57]. The word-based NLP method separately understood word meanings, and sentence meaning was unknown. For example, with the word-based method, a robot understood when a human mentioned the word "cup", but it did not understand the related requests such as "I need a cup of water"[15]. Moreover, this method relied on training samples. If the available training samples are limited, thereby leading to the ignorance of some keywords, the meaning understanding will be poor [51]. These two drawbacks limited robots' understanding to a shallow level where only symbolic NL expressions were analyzed [58][59].

Compared with the word-based understanding method, which enabled a shallow-level understanding, a concept-based method allowed a comprehensive understanding of NL expressions. The concept-based method modeled the meanings for sentences by exploring the embedded concepts, which mainly included implicit NL indications [60][61], hierarchical ontologies [62][63], and semantic correlations [64][65]. The concept-based method not only understood explicit facts such as involvements of action/object/events/persons, but also understood the implicit indications such as action purpose, object usage, event meaning, and human intentions. The concept-based method was widely used in research such as [5][6][23] for complicated NL expression analysis. Compared with the understanding supported by the keyword-based method, the concept-based method endowed a robot with a relatively in-depth understanding of the meanings in NL expressions. However, due to the limited consideration of implicit logic correlations in NL expressions, the knowledge represented by the concept-based method cannot model a structural knowledge [5][17][40] This drawback limited robots' understanding towards task procedures and task-world correlations and further limited NLC in practical situations in the real world.

To support a practical implementation of NL understanding, a narrative-based method was developed to create a more sophisticated knowledge representation in a decision-making-focused [68], real-world-aware [69] and human-cognition-imitated manner [70][71]. In this method, the mechanisms of human reasoning and planning, knowledge-to-world mapping, and logic-based human understanding & learning were focused on in the NLP process. Supported by this method, knowledge was practically used in NLC, further improving robot's knowledge scalability and human-robot-cooperation flexibility.

The developments of NLP techniques are shown in Fig. 2, with detailed time labels.

1.3. Pushing force two: development of HRC

In an early period (about 1940s'[72]), humans started to cooperate with robots by using remote controllers, developing an initial HRC, in which the cooperation requirements for action mapping, task goal mapping, and cooperation naturalness/effectiveness were not considered. As the tasks became complicated, both the robot and the human in HRC were assigned with different roles, such as leader-follower and cooperator-cooperator, to perform different parts of a task with considerations of task goal accomplishments, human-robot communications, robot/human statuses and physical/mental capabilities. Compared with HRI, which focuses on general interactions (detailed HRI reviews are in [4][73][74][75]) for physical/mental assistances without task-goal constrains, HRC focuses on specific cooperation for task fulfillment with task-goal constrains such as task planning and adjusting (detailed HRC reviews are [72][76][77]). In this paper, we emphasized on HRC, specifically exploring the state-of-the-art robotic systems

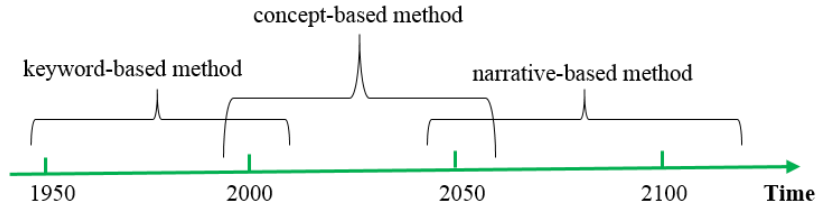

Fig. 2. The development of natural language processing (NLP) techniques [51]. From the stage of keyword-based method, concept-based method, to the narrative-based method, the semantic analysis is becoming sophisticated from word detection to meaning modeling.

using NL to facilitate HRC in various research domains.

Comprehensively, HRC has been developing: it began largely from a low-cognition-level action research where actions were designed and selected according to human instructions, grew to a middle-cognition-level interaction research where shallow-level understanding of human motions, activities, tasks and safety concerns were conducted, and is expanding to a high-cognition-level human-centered engagement research where human's psychological and cognitive statuses, such as attention, motivation, emotion and user models, are considered to improve the HRC effectiveness and naturalness. Increasing human involvements in HRC is shown in Fig. 3.

For action-based HRC, which started from motor-control-based action design, a robot followed simple human instructions to adjust its actions [78][79]. To make the robot actions natural, human-like motion style was then adopted in robot action design [80][81]. Though robots' motion behavior was similar to that of a human, robots' cooperation performances were still poor due to the limited action-understanding being insufficient to support its adaptations towards users/environments [82][83] . To improve robot's adaptability, action interpretations were added for a robot in HRC. For example, 'cup manipulation' in 'drinking' activity meant 'containing liquid'"[40]. Though the understanding was still limited into action level, robots' understanding towards human behaviors in HRC was improved [84]. Overall, action-based HRC was targeting the design of actions in HRC with consideration for cooperation efficiency and naturalness. Due to the lack of human behavior understanding, action-based HRC was still at the imitating level and lacked goal/execution motivation interpretations. Therefore, action-based HRC had a low-level cognition requirement towards robots and could not support an intuitive cooperation.

For interaction-based HRC, which started from action-understanding-based movement imitation [85][86], a robot was required to learn from human demonstrations, understand human movements and develop its own movements. To improve the understanding of human movements, robots were provided with various informative motion data such as human action trajectories [87], hand/body poses [88], and bio-signals [89]. To further improve robots' cooperation performance, robots were trained to explore the mutual influence between a human and his/her surrounding environment [90][91][92][93][94]. Overall, interaction-based HRC improved the cooperation from the naïve predefinition stage to the current intuitive interaction stage. Robot understanding toward human behaviors was improved by using various informative data and considering multiple influential factors from environments. With a comprehensive understanding, a robot was closely associated with a human by correctly identifying the cooperation requests and appropriately providing the assistance.

For engagement-based HRC, individual-level factors, such as individual attentions [95][96], personalities [97][98], emotions [99][100] and safety factors [83][101] , were considered by robots in the cooperation. With the engagement-based HRC method, robot cognition was further improved by adapting various individuals. This method is widely used in present-day HRC research. Given the cognition requirements towards robots in HRC, a natural and efficient cooperation manner such as NLC is in urgent need.

1.4. Systematic overview of NLC research

With cross-disciplinary technique supports, NLC has been developed from low-cognition-level symbol matching control, such as using "yes/no" for controlling robotic execution, to high-cognition-level task understanding, such as "go straight and turn left at the second cross".

As a result of NLC research, a substantial number of projects were launched, including "collaborative research:

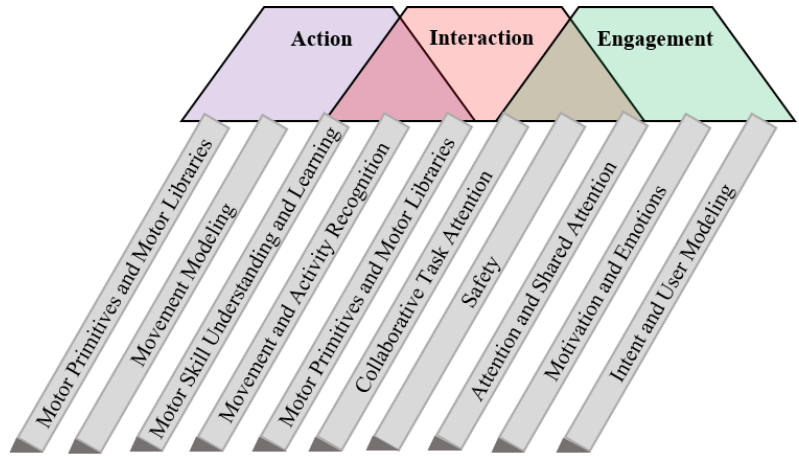

Fig. 3. The human is increasingly involved in HRC [72]. From a perspective of a robot's cognition level, the HRC systems are mainly categorized into action-based HRC, interaction-based HRC and engagement-based HRC. Between the different categories, there are overlaps, showing that the HRC developments happen gradually.

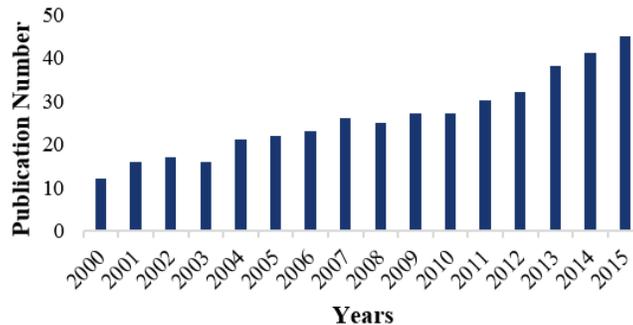

Fig. 4. The annual amount of NLC-related publications since the year 2000 according to our comprehensive paper review. In the past 15 years, the number of NLC publications are steadily increasing and are reaching a history-high level in current time.

jointly learning language and affordances" in Cornell University [102], "robots that learn to communicate with humans through natural dialog" in the University of Texas at Austin [103], "collaborative research: modeling and verification of language-based interaction" in MIT [104], "language grounding in robotics" in University of Washington [105], "semantic systems" in Lund University [106] etc. NLC research is regularly published in international journals such as IJRR [107], TRO [108], AI [109] and KBS [110], and international conferences such as ICRA [111], IROS [112] and AAAI [113]. The publication trend is shown in Fig. 4, where the increasing significance of using NL to facilitate HRC is reflected by the steadily increasing publication numbers in the recent decade.

Compared with the existing review papers about HRC using gesture/pose [114][115], action/motion [116], and tactile [117], a review paper about NLC is lacking. Given the promising potential and increasing attention of NLC, it is necessary to make a summary of the state-of-the-art robotic systems in wide-range domains, revealing the current research progress and signposting future NLC research. The organization of this review paper is shown in Fig. 5.

## 2. Typical systems

As a human-robot-combined decision-making method, the NLC is widely implemented in various robotic systems. According to robot cognition level during the cooperation, NLC-based robotic systems are categorized into four main types: *NL-based control,* where only the NL-format control symbols are given and comprehensive instruction understating is not involved; *NL-based robot training,* where comprehensive instruction understanding is required and intuitive task execution is not conducted; *human-guided task execution,* where comprehensive understanding of human instructions, practical situation conditions and human intentions are required, and intuitive task execution is conducted; and *NL-based social companion*, where the understanding of social norms is required in addition to the NL-based execution conducting. Summary of the typical NLC systems is shown in Table 1.

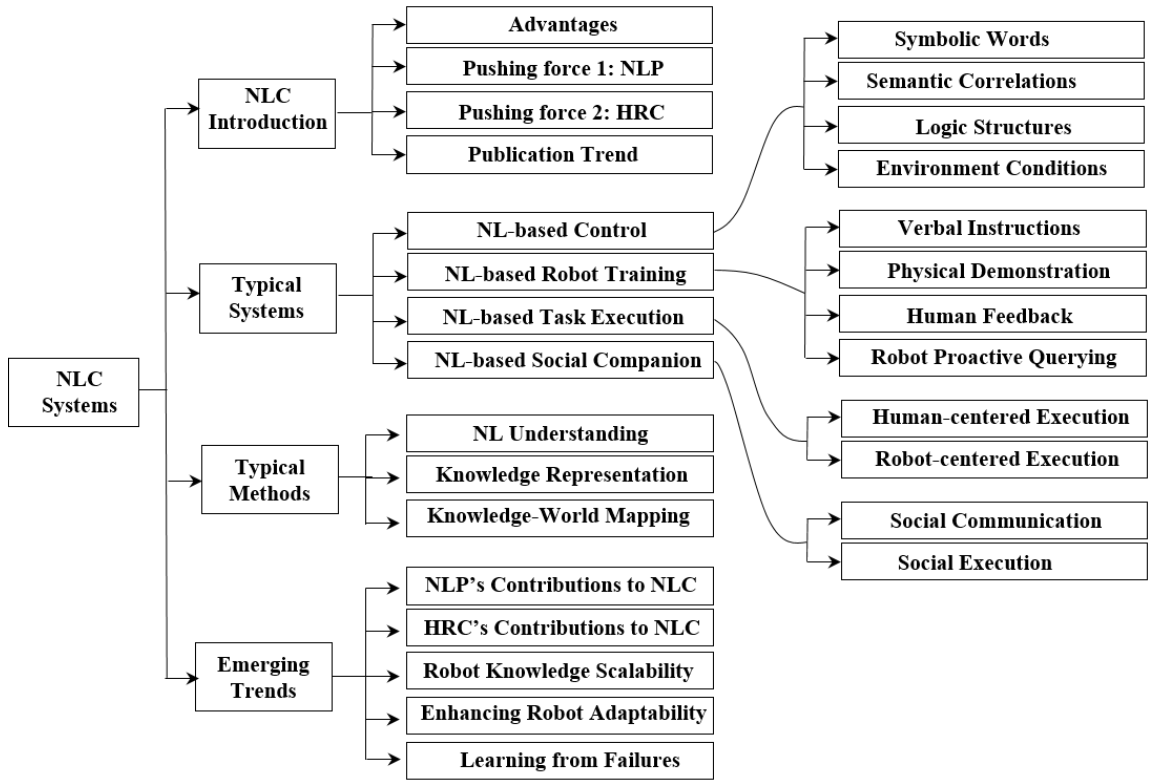

Fig. 5. Organization of this review paper. This review systematically introduced the NLC implementations, covering topics such as the motivations of the NLC research, typical NLC systems, typical NLC methods, and emerging trends. The typical NLC systems were summarized emphatically by categorizing NLC into four categories, including NL-based control, NL-based robot training, NL-based task execution, and NL-based social companion. In each of the categories, the typical application scenarios, knowledge manners/formats, advantages and disadvantages were summarized.

Table 1. Comparison analysis for typical NLC systems

| NLC systems | Application scenarios | Robot cognition level | Human-robot role assignments | Human involvements | Robot involvements |
|---|---|---|---|---|---|
| NL-based control | action selection, manipulation pose adjustment, navigation routine selection | low | leader-follower | all cognitive burden | all physical burden |
| NL-based robot training | execution process explanation (assembly task process, navigation process), action/motion specification (object identification/grasping), human instruction disambiguation (industrial tasks, daily assistance), human-motion imitating | middle | leader-follower | maximal cognitive burden, minimal physical burden | minimal cognitive burden, maximal physical burden |
| NL-based task execution | assembly in dynamic environments with various users, navigation in unstructured environments, heavy object delivering | high | cooperator-cooperator, leader-follower | partial cognitive/physical burden | partial cognitive/physical burden |
| NL-based social cooperation | restaurant reception, social distance maintenance, body language learning during speaking, human-like object manipulation | highest | cooperator-cooperator | partial cognitive/physical burden | partial cognitive/physical burden |

## 3. NL-based control

To set human hands free for other tasks and reduce human's physical burden, NL was initially used to replace physical control means, such as joysticks and remote controllers, which required the use of human hands [33]. During the control, NL played a role as information-delivering media, which contains the human-desired robot executions, shown in Fig. 6. A human user mainly made decisions in the control process, while a robot was controlled by detecting and following the controlling commands in human NL requests. From a burden-assessment perspective, during the whole cooperation process, the human mainly took the cognitive burdens, while the robot mainly took the physical burdens.

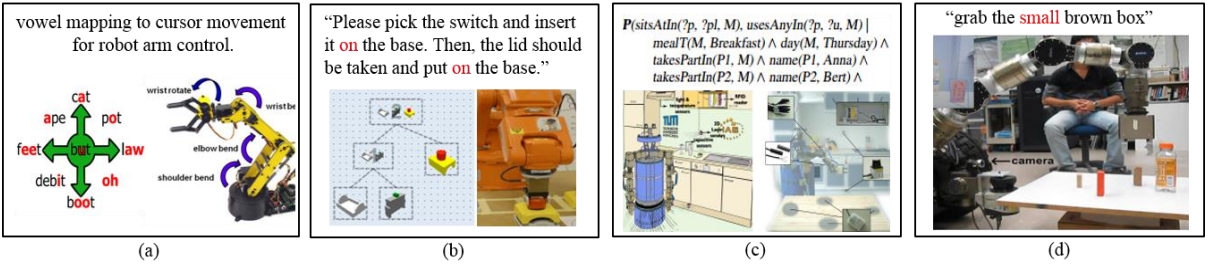

(a) (b) (c) (d)

Fig. 6. Typical robotic systems using NL-based control. (a) Control using symbolic words. The joints' motion directions of a robot arm were controlled by mapping the symbolic vowels from human's oral instructions [128]. (b). Control using semantic correlations. By mapping the semantic correlations of the parts from human's oral instructions, a robot was orally controlled by a human to perform assembling works [129]. (c). Control using logic structures. The cooking task executions were defined in a logic manner. By mapping these logics from human oral instructions, robot performed kitchen tasks [130]. (d). Control using environment conditions. During the task execution, a robot considered the practical environmental conditions such as "object availability, objects' relative sizes" to perform the tasks such as "grab the small brown box" [33].

3.1. Typical robotic systems

*Control using symbolic words.* The idea of using human NL for robot control was proposed at the year 1992. NL was initially used in robot teleoperation [118], in which the correct robot actions were selected by a human to guide it to a desired destination. The system in paper [119] used NL instructions to plan task-execution procedures for a robot. The NL-based control for execution procedure planning was a word-to-action mapping process, which was discrete and the word-action associations were predefined in the robot's database. In this process, a human was restricted to give symbolic and simple NL commands to a robot. The robot needed to accurately detect the symbolic words in human speech and then associate them with the predefined actions or action sequences. Typical NLC systems using symbolic word control include manipulation control [119][120], motion trajectory control [121][122], navigation location & behavior control [88][123]etc. These works addressed the challenges in speech recognition, word disambiguation and word-action mapping. NL-based control was used to specify the detailed action-related parameters such as action direction, movement amplitude, motion speed, force intensity, and hand pose status. The NL-based control for action specification was the word-to-value mapping process, which was continuous and its value ranges were predefined in robots' database. During the development of NL-based control, the mapping rules such as fuzzy [124]/strict [118] mappings were designed.

*Control using* semantic *correlations.* To further improve the robustness of NL-based control during HRC, semantic correlations among the controlling symbols were explored by analyzing the linguistic structures of the NL commands. Controlling symbols include different types of actions, hand poses, objects etc. By exploring the semantic correlations, such as "grasp-cup" and "go-To-left" among these symbols [125], human commonsense was initially involved in robot control process, increasing the naturalness of the robot executions in scenarios such as navigations and manipulations. Moreover, the execution flexibility of a robot was also improved by extracting the general cooperation patterns such as "grasp(object)" [119][126] and "goTo(Location)" [2][127], improving robots' adaptability towards instruction variety.

*Control using logic structures.* Even though semantic correlation improved the flexibility of NL-execution mapping, the control performances in dynamic situations was still limited due to the ignorance of control logics among these semantic correlations. The logic ignorance made a robot incapable of adjusting itself to environmental changes and incapable of intuitively reasoning about the execution plans. For example, the NL instruction "fill the cup with water, deliver it to human" includes logics "search cup, then use water pot, last deliver cup". The ignorance of the logics in the controlling process will lead to the wrong executions such as "use water pot, then search cup" or will restrict the correct executions such as "deliver cup, then use water pot" in dynamic environments, which will cause the removing/adding execution steps. To solve this problem, the logic correlations, including temporal logic, spatial logic and ontology logic among the controlling symbols, were explored to enhance the adaptability of the NL-based control method. Typical robotic systems using logic to facilitate robot control include modifying robot movement trajectory in different situations [127][131], designing robot manipulation post according to fuzzy action type/speed requirements [88][124], serving meals by considering "foodType–vesselShape" relations [130][132], and assembling industrial parts by considering the spatial matching relations [129][133] etc.

*Control using environmental conditions.* Besides the logic relations among the control commands, it is necessary to consider environmental conditions for a practical NL-based control in real-world situations. The balance among human commands, robot knowledge and environment conditions should be made for an intuitive NLC. With the NLC systems, typical applications that consider practical environment conditions include grasping with considering constraints in safety, temporal relations and human preferences [58][134], navigating by considering location/building matching [20][135], serving food with consideration of user locations [136], food vessel shapes and path conditions [119] etc. In these systems, if NL commands and robot knowledge were supported by real world conditions, NLC tasks are highly likely to be successful, or NLC tasks are likely to fail.

3.2 Open problems

For the NL-based robot control, a human interacted with a robot in a verbal manner. Simple NL control commands were given separately, or in a hybrid manner where the NL commands were combined with visual/haptic cues. A human was the only information source, guiding the whole control process. A robot was designed to simply map the human NL commands to the knowledge structure in robot databases, or to the real-world conditions perceived by robots' sensors. With physical/mental work assignments for robots and humans, current efforts in NL-based control focus on improving control accuracy, decreasing humans' cognition burdens and increasing a robot's cognition burdens. However, the cognition burden of humans in NL-based control was still at a high level and robot cognition was still at a low level. A human user was required to lead the cooperation and the robot was required to follow the human with an understanding of control commands. The big cognition-level difference between a human and a robot restrained the intuitiveness and naturalness of the current NLC systems in the cooperation. Summary of NL-based control systems is shown in Table 2.

Table 2. Summary of NL-based control systems

|  | **Control using symbolic words** | **Control using semantic correlations** | **Control using logic structures** | **Control using environment conditions** |
|---|---|---|---|---|
| **Knowledge-providing manner** | predefined | predefined | predefined | predefined + sensing |
| **Knowledge format** | symbolic words "yes, no, go, stop" | linguistic structure | control logic formulas | verbal commands + real-world conditions. |
| **Typical applications** | object manipulation, motion trajectory control, navigation control | grasping, navigation | sequential action control, trajectory modification, hand-pose selection, object recommendation, assembly | safety concerns in grasping, human-preferred temporal action sequence, precise navigation, user-centered meal service |
| **Advantages** | concise, accurate | flexible | flexible, adaptable | flexible, adaptable, consider real-world conditions |
| **Disadvantages** | inflexible, unnatural, limited adaptability | unnatural, limited adaptability on dynamic situations | implicit, unsafe/risky by ignoring real-world conditions | not natural and intelligent for lacking a meaningful interpretation |
| **Typical references** | [120][121][122] | [125][126][127] | [129][130][131] | [134][135][136] |

## 4. NL-based robot training

To support a robot in complex task planning and intuitive decision making, NL was used to train robots for task executions with a spoken/written manner. During the training, knowledge was transferred from robot/human experts to targeted robots, shown in Fig. 7. With consideration of robot physical capabilities such as force/strength/physical structure/speed, human preferences such as motion/emotion, and real-world conditions such as object availability/distribution/location, executable knowledge was specified for a robot. With the executable knowledge, robots' capabilities in task understanding, environment interpretation and human-request reasoning are improved. Different from NL-based control, where a robot was not involving in advanced reasoning, in NL-based robot training, robots were required to reason about human cooperation requests from NL instructions. According to the knowledge transferring manners, systems using NL-based robot training are mainly categorized into four types: *training using*

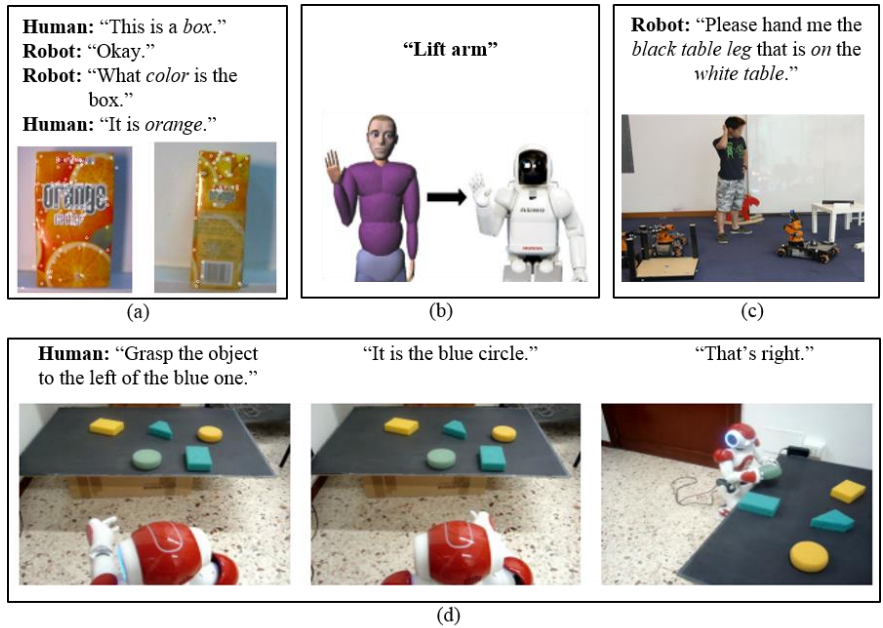

Fig. 7. Typical robotic systems using NL-based robot training. (a) Training using human instructions. By describing the physical properties of the objects, the robot-related knowledge was transferred from a human to a robot, enabling the robot to recognize objects in the future [143]. (b). Training using human demonstrations. With a human's physical demonstrations of actions, a robot learned to perform an action by imitating the motion patterns such as trajectory, action sequence, and motion speed. [150]. (c). Training using proactive robot querying. A robot proactively detected its missing knowledge and proactively ask human for knowledge support [40]. (d). Training using human feedbacks. During the execution process, a human proactively interfered with the execution and gives the timely feedbacks to improve a robot's performances [153].

*human instruction*, *training using human demonstration*, *training using human feedback* and *training using proactive robot querying*. In both the training using human instruction and human demonstration, robots passively learn from a human. The difference is that in instruction humans only orally describe and do not physically participate, while in demonstration, humans need to physically participate. In the training both using human feedback and proactive robot querying, robots proactively learn from a human. The difference is that in the training using human feedback a human decides what knowledge to learn, while in the training using robot querying a robot decides what knowledge to learn.

4.1. Typical robotic systems

*Training using human instruction.* Given that human NL instructions are informative and natural, robot training initially started with using NL instructions to define task execution methods [137]. In the early stage of instruction training, these instructions were used to perform low-level knowledge grounding, in which short NL expressions given by humans were mapped into separated knowledge entities in the real-world. With the NLC systems, typical applications include using the NL instructions to identify an object [132][138], object physical properties [139][140], and action associations [141][142]. Low-level knowledge grounding endowed robots with a shallow understanding of the motivations and logics in the instructed task-execution procedures due to which the knowledge was mapped from a database to the real world in a point-to-point manner. Instead of correlating the execution procedures to form a semantic network for comprehensive execution understanding, the knowledge correlations in the low-level knowledge grounding were mutually independent. As information/automation techniques improved, the low-level knowledge grounding method was then evolved into the high-level knowledge grounding method, in which complex NL expressions were grounded into hierarchical knowledge structures for motivation/logic understanding. With the hierarchical knowledge, the instruction-based training for sophisticated task execution was enabled. With the NLC systems, typical applications include using the NL instructions to describe daily common sense such as precise object manipulation [143][144], to describe spatial/temporal correlations for intuitive navigations [145][146], and to model human cognition for daily activity performing [147][148] . During the high-level knowledge grounding, human cognition processes on task planning and performing were modeled. The advantage of the training using human NL instructions is that robots' reasoning mechanisms during NLC is initially developed; the disadvantage is that the learned execution methods are still abstract in that the sensor-value-level specifications for the NL commands are lacking, limiting the knowledge implementations in real-world situations.

*Training using human demonstration.* To improve robot training in practical perceiving and acting, a human demonstration method was developed to train a robot in real task-execution situations. With training in a human-demonstration manner, theoretical knowledge such as actions, action sequences and object weight/shape/color was associated with sensor data and data patterns. This theory-practice association enabled a straightforward, sensor-data-based interpretation towards the abstract task-related knowledge, improving robots' understanding and cooperation by practically implementing the learned knowledge. A general demonstration process was that a human physically performs a task and meanwhile orally explained the execution intuitions for a robot. The robot was expected to associate the NL-extracted knowledge with sensor data to specify the task executions. With the NLC systems, typical applications using the demonstration-based robot training method include: learning object-manipulation methods by associating human NL expressions with sensor data such as touching force values, object color/shape/size and visual trajectory [144][149]; learning human-like gestures by associating human hand gesture with speech context [150][151]; learning object functional usages by simultaneously considering human voice behaviors, motion behaviors and environment conditions [147]; learning abstract interpretations of environmental conditions by combining human operations, human NL explanations, and the corresponding sensor data patterns [21][151]; adapting new situations by replacing NL-instructed knowledge with real-world-available knowledge [36][152] etc. Human demonstration enabled a robot with a practical understanding of real-world task executions. Compared with robot training using instructions, robot training using demonstrations specified the abstract theoretical knowledge with the real-world, making the learned knowledge executable in real-world situations. However, robots' reasoning capability was not largely improved since demonstration-based training was actually a sensor-data-level imitation of human behaviors and ignored the "unobservable human behaviors" such as human's subjective interpretation towards real-world conditions, human's philosophy in execution and human's cognitive process in decision making.

*Training using human feedback.* To improve robots' reasoning capability in NLC, human NL feedback was used to directly tell the robots "the unobservable human behaviors". With human NL feedback, robot behaviors in cooperation were logically modified by adding/removing some operation steps [40][96][153] or subjectively emphasizing on executions [154][155][156]. With the NLC systems, typical applications include: indicating the human-desired locations/objects with NL instructions during HRC [153]; assessing robot execution performances and correcting robots' undesired manipulation behaviors such as hand poses and object selections by using real-time NL instructions [96]; analyzing robots' execution failures and helping a robot to learn from failures with NL conversations [40]; and emphasizing/ignoring robot behaviors in complex task execution by subjective NL rewards and punishments such as "joy, anger" [155]. Compared with training using human demonstrations, training using human feedback proactively and explicitly indicated a robot with operation logics and decision-making mechanisms, enabling better robot understanding towards human cognitive process in cooperation. Based on both human cognition understanding and environment perceiving, robots' surrounding environments in NLC were interpreted as a human-centered situation. In this human-centered situation, task cooperation was interpreted in a human perspective, improving robots' reasoning capability in cooperating with a human. However, the feedback-based learning required frequent human involvements, imposing a heavy cognition burden on a human. Moreover, the knowledge learned by human feedback was given by a human without considering the robot's actual knowledge needs, limiting robots' adaptation to new environments where robots' knowledge shortage was waiting to be compensated for successful NLC.

*Training using proactive querying.* To solve the new situation adaptation problem for further improving a robot's reasoning ability, a proactive-querying method was developed for robot training. In the querying process, a robot used NL to proactively query humans about its missing knowledge about human-intention disambiguation, environment interpretations, and knowledge-to-world mapping. After the training, a robot was endowed with a more targeted knowledge to adapt the previously-encountered situations, thereby improving a robot's environment adaptability. With the NLC systems, typical applications using querying-based training include: asking for cognitive decisions on trajectory/action/pose selections in tasks such as "human-robot jointly carrying a bulky bumper" [157]; asking for knowledge disambiguation of human commands such as confirming the human-attended object "the blue cup" [158]; asking for human physical assistances to deliver a missing objects or to execute robot-incapable actions such as "deliver a table leg for a robot" [40]; asking for additional information such as "the object is yellow and rectangle" from a human to assisting robots' perceiving [153][159] etc. Compared with the training using human instructions, demonstrations and feedback, in which a robot was trained passively, in the querying method a robot was trained proactively and knowledge was more situation-specific. Robots were endowed with an advanced self-improving capability. Supported by a never-ending mechanism, robot performances in NLC were improved in the long term by continuous knowledge acquiring and refining [91].

4.2. Open problems

During the development of training methods starting from instruction training to querying training, human cognition burden was gradually decreased and robot cognition level was gradually improved. Robot trainings using the above-mentioned methods are suffering the shortcomings of robot knowledge scalability and adaptability. The knowledge scalability problem is caused by limited knowledge from limited information sources such as available humans/robots experts. The robot knowledge is hard to be largely scaled up with an economic manner. Knowledge adaptability problems are caused by the current shallow cognition models. It is hard for a robot to adapt to a specific user/situation and meanwhile adapt to the general types of users/situations. The potential solutions for the knowledge scalability problem and adaptability problem will be discussed in section 8.3 and 8.4. A summary of NL-based robot training systems is shown in Table 3.

Table 3. Summary of the NL-based robot training systems

|  | **Training using human instruction** | **Training using human demonstration** | **Training using human feedbacks** | **Training using proactive query** |
|---|---|---|---|---|
| **Knowledge transferring manner** | human oral descriptions | human physical demonstration | human verb/physical interference | robot verbal querying |
| **Knowledge format** | speech | speech, motion | speech, motion | speech |
| **Human involvement** | instructor | demonstrator | leader | assistant |
| **Robot involvements** | follower | follower | assistant | leader |
| **Typical applications** | object recognition methods, object manipulation methods, task execution procedures | NL-features (force/color/size/visual) association, human-like gesture learning, motion-based object usage learning, environment condition interpretation, abstract human plan execution | human-preferred execution learning, human-like manipulation learning, robot learns from failures | human-in-the-loop decision making, human physical assistance, human intention disambiguation, human-robot knowledge transferring |
| **Advantages** | completed property/process definition | make abstract and ambiguous NL instructions explicit and machine-executable. | human preference consideration, initial human-like cognitive process modeling | robot get knowledge in need, relatively strong environment adaptability |
| **Disadvantages** | ignoring practical environment conditions, limited user/environment adaptability | weak reasoning ability | frequently bother human, heavy human cognitive burden | frequently bother human, heavy human cognitive burden |
| **Typical references** | [139][140][141] | [36][151][152] | [154][155][156] | [157][158][159] |

5. **NL-based task execution**

Different from NL-based robot training systems, in which human NL was helping a robot with its task understanding, in NL-based robot task execution systems, human NL was helping a robot with its task execution. In NL-based training, a robot created a structure-completed and execution-specified knowledge representation. But in NL-based task execution, including understanding the task, the robot was also required to interpret the surrounding environments, predict human intentions, and make optimal decisions satisfying the constraints from the environment, the task execution method and the human requirements. Typical robotic systems using NL-based task execution are shown in Fig. 8. Given that the reasoning was strictly requested in NL-based task execution, robots' cognition levels in NL-based task execution were higher than that in NL-based robot training. With respect to whom is leading the execution, systems of NL-based execution are categorized into human-centered task execution, in which a human is mentally leading the task executions and a robot is providing appropriate assistances for facilitating human executions, and robot-centered task execution, where a robot is mentally leading the task executions and the human is providing oral reminders or physical assistances for facilitating robot executions.

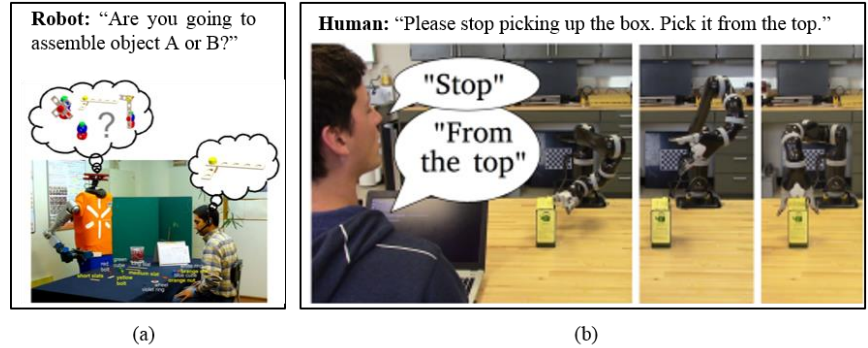

Fig. 8. Typical robotic systems using NL-based task execution. (a) is human-centered execution. A human was performing tasks such as "assemble a toy". A robot was standing by and meanwhile prepares to provide help, ensuring the success and smoothness of the human's task executions. A robot was expected to infer the human's ongoing activities, detect human needs timely and proactively provide the appropriate help such as "a toy part" [160]. (b). Robot-centered execution. A robot was autonomously performing a task. A human was standing by to monitor the robot executions. If abnormal executions or execution failures occurred, the human provided the timely verbal corrections such as "stop, grasp the top" or physical assistances such as "delivering the robot-needed object" [161].

5.1. Typical robotic systems

*Human-centered task execution.* Given the technology limitations in perceiving, reasoning and acting, a robot is still not fully automated and human intelligence is still necessary in HRC. To integrate human mental intelligence and robot physical execution, NL-based execution systems were designed to be human-centered. NL expressions in task execution deliver information such as explanation of human's cooperation requests, descriptions of human's execution plans and indications of human's urgent needs. With this information, a robot provides appropriate assistances timely. With NLC systems, typical applications using human-centered executions include performing tasks such as "table assembly", during which the human sets up task goal (assembly a specific part), makes plans (action steps, pose and tool usages) and partially executes tasks (assemble the parts together), and the robot provides human-desired assistances (tool delivery, part delivery, part holding) [162][163]. During the cooperation, human took both cognitive and physical responsibilities, and the robot took partial physical responsibilities. Comprehensive human-centered execution was developed so that a human was only burdened with cognitive responsibilities such as explaining the navigation routine [164][165], describing the needed objects and location/pose [66][166] and guiding the fine/rough processing [5][167]. Correspondingly, a robot took on only physical responsibilities such as grasping/transferring the fragile/heavy objects [168][169]. Both the human and the robot performed independent sub-tasks by considering the same high-level task goal. The robot received less instructions for its tasks and meanwhile was expected to monitor human's task processes so that the robot provided appropriate assistances when the human required it. This cooperation proposed a relatively-high standard towards robot cognition on providing appropriate assistances at the right location/time. Overall, in human-centered NL-based task execution, the human was leading the execution at the cognition level, and a robot provided the appropriate assistance for saving the human's time and energy, thereby enhancing the human's physical capability.

*Robot-centered task execution.* To further improve the cooperation intuitiveness and decrease human's mental/physical burdens, systems using robot-centered task execution were developed. Different from human-centered systems in which a human mainly took the cognitive/physical burdens and a robot gave human-needed assistances to facilitate human execution, in robot-centered systems, a robot mainly took the cognitive/physical burdens, and a human gives robot-needed assistance physically to facilitate the robot execution. NL expressions in the robot-centered systems were used for a robot to ask for assistances from a human. With the NLC systems, typical applications using robot-centered task execution include: robot lead industrial assembly, in which a human enhanced robot physical capability by providing robot with physical assistances such as grasping [170] and fetching [171]; robot executed tasks such as heavy object moving and elderly navigation in unstructured outdoor environments, in which a human analyzed and conquered the environment limitations such as objects/space availability [22][172][173][174]. Compared with robots in human-centered execution, where a robot was required to comprehensively understand human behaviors, robots in robot-centered applications were required to comprehensively understand the limitations on robot knowledge, real-world conditions, both humans' and robots' physical capabilities etc. The advantage was that the human was less involved and his/her hands/mind were partially set free.

5.2. Open problems

An open problem for human-centered task execution is that the human's cognition burden is relatively heavy. Even though a robot become more autonomous to prove timely assistances, a human is still leading the NLC mentally/physically. An open problem for robot-centered task execution is that robot-centered applications set a high standard towards the robot capability on situation analyzing, plan making, assistance asking, and knowledge updating. While these robot capabilities are still immature [175]. Therefore, a NLC-based system involving human cognitive process modeling, intelligent robot decision-making, autonomous robotic task-execution, and human/robot physical capability consideration is in urgent need. Summary of NL-based task execution systems is shown in Table 4.

Table 4. Summary of the NL-based task execution systems

|  | Human-centered task execution | Robot-centered task execution |
| --- | --- | --- |
| **Activity recognition requirement** | robot, high | human, high |
| **Human involvements** | high | low |
| **Human's job** | cognitive burden | physical burden |
| **Robot's job** | physical burden | cognitive burden |
| **Human involvement** | leader | assistant |
| **Robot involvements** | assistant | leader |
| **Typical applications** | assembly, object grasping | daily assistance, industrial assembly, outdoor navigation, heavy object delivering |
| **Advantages** | accurately assistance | strong environment adaptation |
| **Disadvantages** | heavy cognitive burdens on human, high requirements on robot' reasoning capability on recognizing human activity and detecting human needs | human cognitive burden, human disturbance |
| **Typical references** | [167][168][169] | [172][173][174] |

## 6. NL-based social cooperation

To make NLC natural and intuitive, NL-based social cooperation was developed by involving human's social behaviors such as social speeches and social motions in NLC. Different from NL-based task execution systems, which focused on objective task information such as task goals and task execution procedures, NL-based social cooperation systems focused on subjective social norms such as social speech or motion behaviors, shown in Fig. 9. With social norms, robots were naturally integrated with a human in NLC and, meanwhile, robots' social acceptances were increased. According to social norm types, robotic systems using NL-based social cooperation are categorized into *social communication* in which social NL expressions are used for facilitating communication, and *social execution* in which social NL expressions are used for facilitating execution.

6.1. Typical robotic systems

*NL-based social communication*. In a natural and intuitive NLC, robot-human communication needs to be both information-correct and social-norm appropriate. Capturing social norms from humans' NL expressions was helpful in certain aspects such as detecting human preferences in cooperation, specifying cooperation roles such as "leader, follower, cooperator", and increasing social acceptance. NL in social communications served as an information source, from which both the objective execution methods and the subjective human preferences were extracted. With the NLC systems, typical applications using NL-based social communication include: a receptionist robot increased its social acceptance in conference arrangements by using social dialogs with pleasant prosodic contours [180]; cooperative machine operations used social descriptions considering human action preferences [181][182]; health-caregiving robots searched and delivered objects by considering user speech confidences, user safety and user roles such as "primary user, bystander" [183]; adapted unfamiliar users by using NL expressions with fuzzy emotion statuses such as "fuzzy happiness/sadness/anger" [184]; modeled social NL communications in NLC by defining human-robot relations such as "love", "friendship" and "marriage" [185]; designed robotic companion by using friendly NL conversation [67][186][187] etc.

*NL-based social execution.* In a natural and intuitive NLC, the communication and the execution need to be socially appropriate. To introduce social execution norms into NLC, NL-based social execution systems were designed. NL was used to indicate socially-preferred executions for robots, enhancing robots' understanding of social motivations behind task executions and further making robot executions socially-acceptable. With the NLC systems, typical applications using NL-based social executions include: a navigation robot autonomously modified its motion behaviors (stop, slower, faster) by considering human density (crowded, dense) with the reminding of human NL instructions ("go ahead to move", "stop") [188]; a companion robot moved its head towards the human speaker according to human's NL tunes [189]; a storytelling robot depictd stories by mapping NL expressions with human's body motion behaviors to catch humans' attention [178][190].

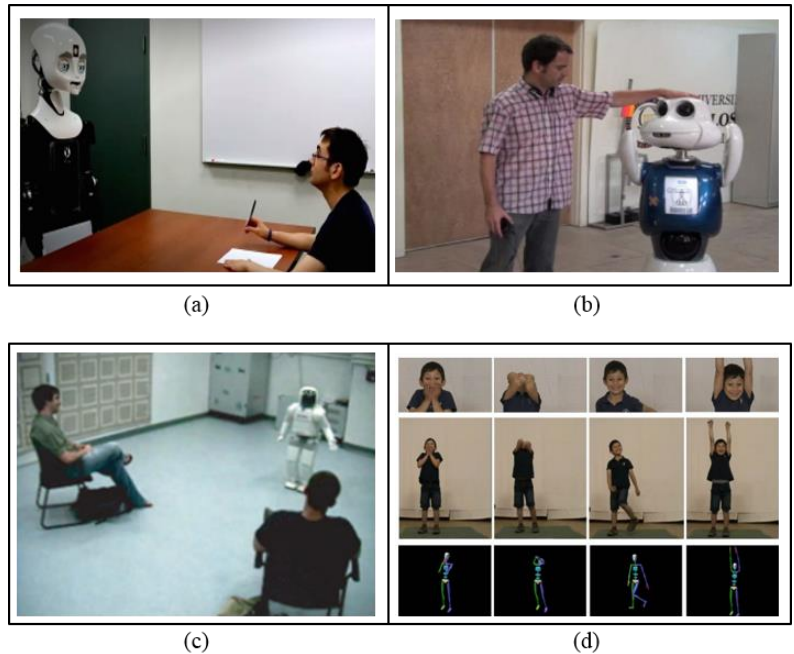

Fig. 9. Typical robotic systems using NL-based social cooperation. (a) and (b) are NL-based social communication. A robot learned to nicely response a human's request such as "drawing a picture on the paper" [176] and "stop until I touch you" [177]. (c) and (d) are NL-based social execution. A robot learned to do appropriate body languages during its speaking such as storytelling [178][179].

NL-based social communication and NL-based social execution focused on two different aspects of NLC. To develop socially intuitive NLC systems, the two aspects need to be focused on simultaneously.

6.2. Open problems

Social norms in both communications and executions are hard to model. First, social norms are *implicit*. It is challenging to summarize social norms from human behaviors. Second, social norms *vary*. Different regions, countries, cultures, races and personalities form different social norms. A social-norm model that considers the above influential factors is challenging to create because the representative norms are difficult to extract. Last, social norms are currently *non-evaluable*. It is challenging to assess the correctness of social norms because there are no clear standards to judge the correctness of social norms, and different persons have different levels of social behavior acceptance/tolerance degree. Summary of NL-based social cooperation systems is shown in Table 5.

Table 5. Summary of NL-based social cooperation systems

|  | **NL-based social communication** | **NL-based social execution** |
|---|---|---|
| **Knowledge transferring manner** | verbal | physical |
| **Knowledge format** | NL expressions | physical execution |
| **Human involvement** | both cognitive & physical burdens | both cognitive & physical burdens |
| **Robot involvements** | both cognitive & physical burdens | both cognitive & physical burdens |
| **Typical applications** | restaurant receptionist, health caregiving, industrial cooperation, human cooperation | social distance maintenance, body language learning in storytelling, social behavior-supported object carriage, object manipulation using social expressions |
| **Advantages** | more social acceptance, individualized | understand/respect human's social preferences, understand/respect human expectations |
| **Disadvantages** | hard to model user-specific social expressions | hard to model user-specific social execution manners. |
| **Typical references** | [181][182][183] | [188][189][190] |

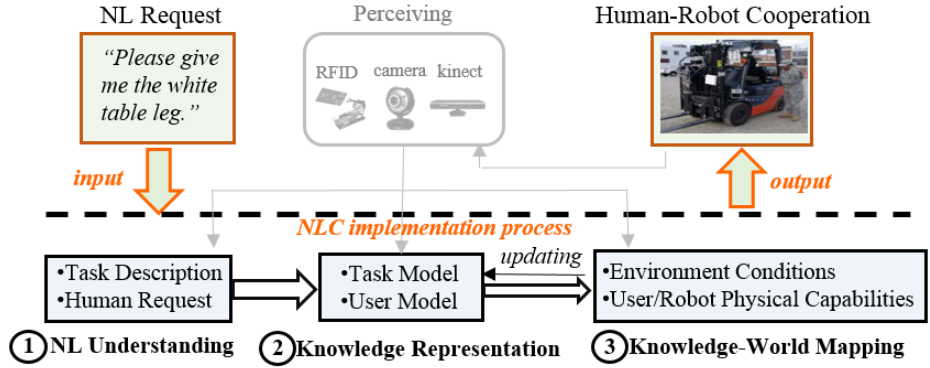

Fig. 10. Main steps for realizing NLC. The input for a NLC method is the human NL request, the output is the HRC [191]. In the NLC implementation process, there are mainly three steps, including NL understanding in which the task-related knowledge is extracted from the human NL requests, knowledge representation in which the extracted knowledge is organized in an algorithm structure, and knowledge-world mapping where theoretical knowledge is grounded into real-world situations.

## 7. Methods for realizing NLC

Developing NLC systems are suffering from three main challenges. First, it is challenging to comprehensively understand human NL instructions. Understanding NL instructions is not about precise speech recognition, but instead precise semantic analysis, by which the meaning, logic and human cognitive process embedded in NL are extracted. Second, it is challenging to represent robots' understanding to support robots' decision-making in NLC. The representation of a robot's understanding is expected to include the task-related knowledge contents and knowledge implementation manners. With the representations, a robot is able to measure the applicability of its knowledge in a situation and decide the correct contents and manners of using its knowledge. Third, it is challenging to accurately map a robot's knowledge into real-world situations. In the mapping process, the theoretical items in knowledge databases are expected to be associated with the corresponding practical objects or relations in the real world, and the incomplete/real-world-inconsistent knowledge is expected to be autonomously corrected.

To address these three challenges, three types of research were conducted: NL understanding, knowledge representation, and knowledge-world mapping. With the NL understanding research, human verbal instructions were processed and the task-related knowledge, such as task goal, execution steps, and the execution parameters "speed, tools, locations, human requirements" were extracted to perform a comprehensive semantic analysis [5][172][192] . With the knowledge representation research, task knowledge was constructed by algorithms such as Bayesian Network [91][193], Support Vector Machine (SVM) [194][195], and Hidden Markov Model (HMM) [118][196]. By using these algorithm structures, the knowledge was capable of supporting robot decision making with various logic/semantic/spatial/temporal manners. With the knowledge-world mapping research, the theoretical information patterns were grounded into real-world situations correctly, and the incomplete/inconsistent knowledge gaps were filled to ensure the success of NLC. In theoretical knowledge grounding, the grounding methods aimed to explore the semantic correlations among the theoretical knowledge and the real-world by associating the temporal/spatial/visual/physical features [197][198] . In the gap filling, the methods aimed to detect the missing knowledge, which was needed in real-world situations but ignored in theoretical knowledge representations, and the inconsistent knowledge, which was instructed by a human but was not available in the real world [58][199] [200]. The complete NLC realization process by using the above-mentioned researches is shown in Fig. 10.

## 8. Emerging trends

NLC has been developed to improve the effectiveness and naturalness of HRC. Due to the limitations of NLC-related techniques such as NLP, machine learning and robot design, NLC performances in dealing with complex tasks, various users, and dynamic/unstructured environments still needs to be improved. Based on our comprehensive review, the future trends for future NLC research are summarized as follows.

8.1. NLP's contributions to NLC

Comprehensively, NLP is undergoing a deep-neural-network revolution to create sophisticated computational sematic models, including the word embedding method for comprehensive meaning modeling by adding in extra meanings such as "cats and dogs are animals" in data preparation stages [201], sequence-to-sequence language understanding/translation by sequentially outputting the meaning-modeling results based on the previous semantic context [202], attention-based NL understanding in which the relatively important words/expressions are valued by increasing the weights of the important expressions and decreasing the weights of the unimportant expressions [203], unsupervised long-term meaning modeling [204] etc.

With the advantages brought by the NLP revolution, NLC is correspondingly benefited in areas including: complex cooperation request understanding by adding task-specific knowledge such as manufacturing common sense; daily assistance common sense and caregiving common sense into the input NL data; real-time and context-aware task execution by aligning NL expressions with knowledge related to tasks, robots, environments, and humans; human-desired execution priority analysis by analyzing human's verbal focuses in communication; self-improving of robot understanding ability during task executions.

8.2. HRC's contributions to NLC

The current HRC research has two trends: *generalization* and *specification*. In the generalization trend, a robot is endowed with broad commonsense knowledge to support the general NLC under various situations [33][94][205]. Massive commonsense is essentially the general principle during typical task executions. The objective of generalization is to make a robot adapt to a wide range of tasks/situations/users. To effectively summarize the representative features shared by various NLC scenarios, effective feature learning methods and accurate knowledge-world mapping models are required. Developing these methods could be a future direction of NLC research. In the specification trend, a robot is endowed with delicate knowledge to support the specific types of HRC [206][207]. The delicate knowledge provides execution details for practical robot execution. The objective is to endow a robot with high professionality on specific tasks in specific environmental conditions with specific users. It is critical for a robot to have a broad and delicate knowledge for intuitive NLC.

Both trends are suffering from challenges. For the generalization trend, NLC emphasizes general situation adaptation by extracting the common-sharing knowledge in various scenarios and ignoring the unique knowledge in individual scenarios. Caused by the unique-knowledge ignorance, general NLC methods are relatively simple, being challenged by cold-start phenomenon, which refers the knowledge not being learned in the training stage and causing execution failures in the implementation stage [208]. For the specification trend, limited by the knowledge coverage range, the robot-capable task types are within a narrow range. A robot, which is expected to execute a specific type of task with high professionality, is incapable of executing a wide range of tasks due to intrinsic mental/physical conflictions between being specific and being general [209].

NL is informative, containing the general execution knowledge such as a typical execution method, and the specific knowledge such as a user's emotions, preferences and personalities. HRC using NL is a useful way to balance the generation and specification trends in robotic research. During NLC, NL transfers the general knowledge and emphasizes the specific knowledge in the cooperation process. A future trend of NLC could be using advanced NLP techniques to realize the mutual compensation between the robot generalization and specification.

8.3. Robot knowledge scalability

Scaling up robot knowledge for supporting robot decision making is a critical issue. On one hand, to understand human NL instructions, represent cooperation tasks, or fill up the knowledge gaps, the effective knowledge scaling-up capability is needed to accurately learn a large amount of knowledge. On the other hand, the time/labor cost are expected to be reduced. Currently, the knowledge-scaling-up research goes in two directions: existed-knowledge exploitation, and new-knowledge exploration. In existed-knowledge exploitation, the abstract meanings of existed knowledge are summarized at a high level to increase the knowledge interchangeability, making one type of knowledge useful in other similar scenarios. The new-knowledge exploration includes: human-based methods, which query humans for new knowledge, and the big-data-based method, which is an automatic and low-cost information retrieval method that extracts knowledge from information sources such as the World Wide Web [205], books [210], machine operation log files [211], and videos [212]. Given the new techniques in computer science and neuro science, there is still a need to develop efficient, low-cost and large-scale knowledge scaling-up methods.

8.4. Enhancing robot adaptability

Weak robot adaptability is typically caused by failures in *execution importance modeling*, based on which the execution priority is made, and *execution interchangeability modeling*, based on which the execution flexibility is made. To increase robot adaptability, new research was launched to model the human cognition process [213][214], which aims to explore humans' decision-making mechanism for modeling robot execution priority and flexibility.

For execution priority, not all the executions are essential for the execution success. For example, in the task "assembly", the procedure "clean the place" is much less important than the procedure "install the screw". For interchangeability, a tool request "deliver me a brush" does not necessarily mean the involvement of a specific tool "brush", but instead means a practical purpose "cleaning the surface" [15]. By knowing these meanings, the cooperation plans are flexibly changed by ignoring the trivial execution procedures and focusing on the important procedures, and replacing the unavailable tools with the other available similar-function tools. Current methods focus on exploring object affordances (object-action correlation) [215], and lacking the in-depth interpretations of task cooperation. In the future, NLC research could be methods that interpret robot executions from a human perspective, improving robot adaptability in unstructured environments and unfamiliar human users.

8.5. Learning from failures

Execution failure causes unnatural task execution or even task failures. The learning-from-failure mechanism has been implemented in computer science for algorithm efficiency improvement [216] and in material science for new material discovery [217]. By exploring information in failure experiences, robots' performance in task execution is improved by avoiding similar failures in the future. In NLC, learning from failure is involved in a definition-based manner [40], in which the failure is analyzed by comparing the available knowledge with the defined knowledge, lacking the analysis of failure causes and recovery mechanism. Therefore, in NLC, learning from failure is also a promising research direction. From our perspective, the potential research problems could be in-depth failure cause analysis, concise NL failure explaining to a human, proactive knowledge updating methods for recovering from the failures, etc.

## 9. Conclusion

This review summarized the state-of-the-art robotic systems for using natural language (NL) to facilitate human-robot cooperation (HRC), thereby providing a summary and comparisons of the natural-language-facilitated human-robot cooperation (NLC) systems. Regarding the robot-cognition levels, NLC systems were categorized into four types: NL-based control, NL-based robot training, NL-based task execution, and NL-based social companion. Based on our perspective and comprehensive paper review, the current emerging trends of NLC research were discussed, providing helpful information for the future of NLC research.

## 10. Acknowledgements

We would like to thank Mr. Xu Zhou, Ms. Natalie Kalin and Mr. Ian Coberly for helping us with work finalization.

## References

[1] J. Baraglia, M. Cakmak, Y. Nagai, R. Rao, and M. Asada, "Initiative in robot assistance during collaborative task execution," in *2016 11th ACM/IEEE International Conference on Human-Robot Interaction (HRI)*, pp. 67-74, 2016.
[2] G. Gemignani, E. Bastianelli, and D. Nardi, "Teaching robots parametrized executable plans through spoken interaction," in *International Conference on Autonomous Agents* and *Multiagent Systems,* pp. 851-859, 2015.
[3] J. Brooks, C. Lignos, C. Finucane, M. S. Medvedev, I. Perera, V. Raman, H. KressGazit, M. Marcus, and H. A. Yanco, "Make it so: continuous, flexible natural language interaction with an autonomous robot," in *Workshops at the 26th AAAI Conference on Artificial Intelligence*, 2012.
[4] T. Fong, C. Thorpe, and C. Baur, "Collaboration, dialogue, and human-robot interaction," in *Proceedings of International Symposium of Robotics Research*, 2001.
[5] R. Liu, J. Webb, and X. Zhang, "Natural-language-instructed industrial task execution," in *ASME International Design Engineering Technical Conferences and Computers and Information in Engineering Conference*, pp. V01BT02A043-V01BT02A04, 2016.
[6] S. A. Tellex, T. F. Kollar, S. R. Dickerson, M. R. Walter, A. Banerjee, S. Teller, and N. Roy, "Understanding natural language commands for robotic navigation and mobile manipulation," *AAAI Conference on Artificial Intelligence*, 2011.
[7] H. Iwata and S. Sugano, "Human-robot-contact-state identification based on tactile recognition," *IEEE Transactions on Industrial Electronics*, vol. 52, no. 6, pp. 1468-1477, 2005.
[8] J. Kruger and D. Surdilovic, "Hand force adjustment: robust control of force-coupled human–robot-interaction in assembly processes," *CIRP Annals - Manufacturing Technology*, vol. 57, no. 1, pp. 41-44, 2008.
[9] E. Colgate and N. Hogan, "An analysis of contact instability in terms of passive physical equivalents," in *IEEE International Conference on Robotics and Automation*, pp. 404-409, 1989.
[10] H. Kjellström, J. Romero, and D. Kragić, "Visual object-action recognition: Inferring object affordances from human demonstration," *Computer Vision and Image Understanding*, vol. 115, no. 1, pp. 81-90, 2011.
[11] S. Kim, J. Jung, S. Kavuri, and M. Lee, "Intention estimation and recommendation system based on attention sharing," in *International Conference on Neural Information Processing (ICONIP)*, pp. 395-402, 2013.
[12] N. Hu, G. Englebienne, Z. Lou, and B. Krose, "Latent hierarchical model for activity recognition," *arxiv: 1503.01820v1*, 2015.
[13] P. Barattini, C. Morand, and N. Robertson, "A proposed gesture set for the control of industrial collaborative robots," in *IEEE International Symposium on Robot and Human Interactive Communication (RO-MAN)*, pp. 132-137, 2012
[14] A. Jain, S. Sharma, T. Joachims, and A. Saxena, "Learning preferences for manipulation tasks from online coactive feedback," *International Journal of Robotics Research*, vol. 34, no. 10, pp. 1296-1313, 2015.
[15] R. Liu and X. Zhang, "Understanding human behaviors with an object functional role perspective for robotics," *IEEE Transactions on Cognitive and Developmental Systems*, vol. 8, no. 2, pp. 115-127, 2016.
[16] K. Ramirez-Amaro, M. Beetz, and G. Cheng, "Transferring skills to humanoid robots by extracting semantic representations from observations of human activities," *Artificial Intelligence*, DOI: http://dx.doi.org/10.1016/j.artint.2015.08.009, 2015.

[17] K. Zampogiannis, Y. Yang, C. Fermuller, and Y. Aloimonos, "Learning the spatial semantics of manipulation actions through preposition grounding," in *IEEE International Conference on Robotics and Automation (ICRA)*, pp. 1389-1396, 2015.
[18] W. Takano and Y. Nakamura, "Action database for categorizing and inferring human poses from video sequence," *Robotics and Autonomous Systems*, vol. 70, pp. 116-125, 2015.
[19] K. Andrej and F. Li, "Deep visual-semantic alignments for generating image descriptions," in *Computer Vision and Pattern Recognition (CVPR)*, pp. 3128-3137, 2015.
[20] V. Raman, C. Lignos, C. Finucane, K. C. T. Lee, M. Marcus, and H. Kress-Gazit, "Sorry Dave, I'm afraid I can't do that: explaining unachievable robot tasks using natural language," in *Robotics: Science and Systems*, vol. 2, no. 1, pp. 2, 2013.
[21] M. R. Walter, S. M. Hemachandra, B. S. Homberg, S. Tellex, and S. Teller, "Learning semantic maps from natural language descriptions," in *Robotics: Science and Systems*, 2013.
[22] F. Duvallet, M. R. Walter, T. Howard, S. Hemachandra, J. Oh, S. Teller, N. Roy, and A. Stentz, "Inferring maps and behaviors from natural language instructions," in *International Symposium on Experimental Robotics*, pp. 373-388, 2014.
[23] C. Matuszek, E. Herbst, L. Zettlemoyer, and D. Fox, "Learning to parse natural language commands to a robot control system," in *13th International Symposium on Experimental Robotics*, pp. 403-415, 2013.
[24] C. Ott, D. Lee, and Y. Nakamura, "Motion capture based human motion recognition and imitation by direct marker control," in *IEEE-RAS International Conference on Humanoid Robots*, pp. 399-405, 2008.
[25] S. Waldherr, R. Romero, and S. Thrun, "A gesture based interface for human-robot interaction," *Autonomous Robots*, vol. 9, no. 2, pp. 151-173, 2000.
[26] R. Dillmann, "Teaching and learning of robot tasks via observation of human performance," *Robotics and Autonomous Systems*, vol. 47, no. 2-3, pp. 109-116, 2004.
[27] K. Sakita, K. Ogawara, S. Murakami, K. Kawamura, and K. Ikeuchi, "Flexible cooperation between human and robot by interpreting human intention from gaze information," in *IEEE/RSJ International Conference on Intelligent Robots and Systems (IROS)*, pp. 846-851, 2004.
[28] Y. Yang, Y. Li, C. Fermuller, and Y. Aloimonos, "Robot learning manipulation action plans by 'watching' unconstrained videos from the world wide web," in *Proceedings of the Twenty-Ninth AAAI Conference on Artificial Intelligence*, pp. 3686-3692, 2015.
[29] J. R. Medina, M. Shelley, D. Lee, W. Takano, and S. Hirche, "Towards interactive physical robotic assistance: parameterizing motion primitives through natural language," in *IEEE International Symposium on Robot and Human Interactive Communication (RO-MAN)*, pp. 1097-1102, 2012.
[30] S. Hemachandra and M. R. Walter, "Information-theoretic dialog to improve spatial-semantic representations" in *IEEE/RSJ International Conference on Intelligent Robots and Systems (IROS)*, pp. 5115-5121, 2015.
[31] S. Hunston and G. Francis, "Pattern grammar: A corpus-driven approach to the lexical grammar of English," *Computational Linguistics*, vol. 27, no. 2, pp. 318-320, 2000.
[32] J. L. Bybee and P. J. Hopper, eds., *Frequency and the emergence of linguistic structure*. Vol. 45. John Benjamins Publishing, 2001.
[33] Y. Cheng, Y. Jia, R. Fang, L. She, N. Xi, and J. Chai, "Modelling and analysis of natural language controlled robotic systems," *IFAC Proceedings Volumes*, vol. 47, no. 3, pp. 11767-11772, 2014.
[34] M. Tenorth, D. Nyga, and M. Beetz, "Understanding and executing instructions for everyday manipulation tasks from the world wide web," in *IEEE International Conference on Robotics and Automation (ICRA)*, pp. 1486-1491, 2010.
[35] C. Wu, I. Lenz, and A. Saxena, "Hierarchical semantic labeling for task-relevant rgb-d perception," in *Robotics: Science and Systems*, 2014.
[36] S. Hemachandra, F. Duvallet, T. M. Howard, N. Roy, A. Stentz, and M. R. Walter, "Learning models for following natural language directions in unknown environments," in *IEEE International Conference on Robotics and Automation (ICRA)*, pp. 5608-5615, 2015.
[37] M. Tenorth, A. Perzylo, R. Lafrenz, and M. Beetz, "The RoboEarth language: representing and exchanging knowledge about actions, objects, and environments," in *IEEE International Conference on Robotics and Automation (ICRA)*, pp. 1284-1289, 2012.
[38] J. Pineau, R. West, A. Atrash, J. Villemure, and F. Routhier, "On the feasibility of using a standardized test for evaluating a speech-controlled smart wheelchair," *International Journal of Intelligent Control and System*, vol. 16, no. 2, pp. 124-131, 2011.
[39] C. Granata, M. Chetouani, A. Tapus and P. Bidaud, "Voice and graphical -based interfaces for interaction with a robot dedicated to elderly and people with cognitive disorders," pp. 785-790, 2010.
[40] R. A. Knepper, S. Tellex, A. Li, N. Roy, and D. Rus, "Recovering from failure by asking for help," *Autonomous Robots*, vol. 39, no. 3, pp. 347-362, 2015.
[41] M. Stenmark and J. Malec, "A helping hand: industrial robotics, knowledge and user-oriented services," in *IEEE/RSJ International Conference on Intelligent Robots and Systems (IROS)*, 2013
[42] R. Schulz, B. Talbot, O. Lam, F. Dayoub, P. Corke, B. Upcroft, and G. Wyeth, "Robot navigation using human cues: a robot navigation system for symbolic goal-directed exploration," in *IEEE International Conference on Robotics and Automation (ICRA)*, pp. 1100-1105, 2015.
[43] A. Boularias, F. Duvallet, J. Oh, and A. Stentz, "Grounding spatial relations for outdoor robot navigation," in *IEEE International Conference on Robotics and Automation (ICRA)*, pp. 1976-1982, 2015.
[44] J. Kory and C. Breazeal, "Storytelling with robots: Learning companions for preschool children's language development," in *IEEE International Symposium on Robot and Human Interactive Communication*, pp. 643-648, 2014.
[45] M. J. Salvador, S. Silver, and M. H. Mahoor, "An emotion recognition comparative study of autistic and typically-developing children using the zeno robot," in *IEEE International Conference on Robotics and Automation (ICRA)*, pp. 6128-6133, 2015.
[46] C. Breazeal, "Social interactions in HRI: The robot view," *IEEE Transactions on Systems, Man, and Cybernetics,* vol. 34, no. 2, pp. 181-186, 2004.
[47] S. B. Kotsiantis, I. Zaharakis, and P. Pintelas, "Supervised machine learning: A review of classification techniques," *Informatica*, vol. 31, no. 3, pp. 249-268, 2007.
[48] P. Berkhin, "A survey of clustering data mining techniques," *Grouping multidimensional data*, vol. 43, no. 1, pp. 25-71, 2006.
[49] S. Ding, H. Zhu, W. Jia, and C. Su, "A survey on feature extraction for pattern recognition," *Artificial Intelligence Review*, vol. 37, no. 3, pp. 169-180, 2012.
[50] N. Ranjan, K. Mundada, and K. Phaltane, "A Survey on Techniques in NLP," International Journal of Computer Applications, vol. 134, no. 8, pp.6-9, 2016.
[51] E. Cambria and B. White, "Jumping NLP curves: A review of natural language processing research," *IEEE Computational Intelligence Magazine*, vol. 9, no. 2, pp. 48-57, 2014.
[52] A. F. Smeaton, "Progress in the application of natural language processing to information retrieval tasks," *Computer journal*, vol. 35, no. 3, pp. 268-278, 1992.
[53] A. Ortony, G. Clore, and A. Collins, *The cognitive structure of emotions*, Cambridge, U.K.: Cambridge Univ. Press, 1988.
[54] N. Bush, "The predictive value of transitional probability for word-boundary palatalization in English," M.S. thesis, Univ. New Mexico, Albuquerque, NM, 1999.
[55] L. Goldin and D. M. Berry, "AbstFinder, a prototype natural language text abstraction finder for use in requirements elicitation," *Automated Software Engineering*, vol. 4, no. 4, pp. 375-412, 1997.
[56] R. K. Taira and S. G. Soderland, "A statistical natural language processor for medical reports," in *Proceedings of the AMIA Symposium*, 1999.
[57] B. B. Rieger, "Semantic relevance and aspect dependency in a given subject domain: contents-driven algorithmic processing of fuzzy word meanings to form dynamic stereotype representations," in *International Conference on Computational Linguistics*, pp. 298-301, 1984.
[58] E. Scioni, G. Borghesan, H. Bruyninckx, and M. Bonfe, "Bridging the gap between discrete symbolic planning and optimization-based robot control," in *IEEE International Conference on Robotics* and *Automation (ICRA)*, pp. 5075-5081, 2015.
[59] N. Dantam and M. Stillman, "The motion grammar: Analysis of a linguistic method for robot control," *IEEE Transactions on Robotics*, vol. 29, no. 3, pp. 704-718, 2013.
[60] L. Denoyer, H. Zaragoza, and P. Gallinari, "HMM-based passage models for document classification and ranking," in *European Colloquium on Information Retrieval Research*, Darmstadt, pp. 126-135, 2001.
[61] T. Hofmann, "Unsupervised learning by probabilistic latent semantic analysis," *Machine Learning*, vol. 42, nos.1–2, pp. 177–196, 2001.
[62] J. E. Busch, A. D. Lin, P. J. Graydon, and M. Caudill, "Ontology-based parser for natural language processing," *U.S. Patent No. 7,027,974*, 2006.

[63] H. Alani, S. Kim, D. E. Millard, M. J. Weal, W. Hall, P. H. Lewis, and N. R. Shadbolt, "Automatic ontology-based knowledge extraction from web documents," *IEEE Intelligent Systems*, vol. 18, no. 1, pp. 14-21, 2003.
[64] E. Cambria and A. Hussain, *Sentic computing: Techniques, tools,* and *applications*. Dordrecht, The Netherlands: Springer-Verlag, 2012.
[65] D. Olsher, "COGVIEW & INTELNET: Nuanced energy-based knowledge representation and integrated cognitive-conceptual framework for realistic culture, values, and concept-affected systems simulation," in *IEEE Symposium on Computational Intelligence for Human-like Intelligence (CIHLI)*, pp. 82–91, 2013.
[66] E. Ovchinnikova, M. Wachter, V. Wittenbeck, and T. Asfour, "Multi-purpose natural language understanding linked to sensorimotor experience in humanoid robots," in *IEEE-RAS 15th International Conference on Humanoid Robots (Humanoids)*, pp. 365-372, 2015.
[67] T. Belpaeme, P. Baxter, J. D. Greeff, J. Kennedy, R. Read, R. Looije, M. Neerincx, I. Baroni, and M. C. Zelati, "Child-robot interaction: Perspectives and challenges," in *International Conference on Social Robotics (ICSR)*, pp. 452-459, 2013.
[68] R. Young, "Story and discourse: A bipartite model of narrative generation in virtual worlds," *Interaction Studies*, vol. 8, no. 2, pp. 177–208, 2007.
[69] E. Mueller, "Modeling space and time in narratives about restaurants," *Literary Linguistic Computing*, vol. 22, no.1, pp. 67–84, 2007.
[70] F. Bex, H. Prakken, and B. Verheij, "Formalizing argumentative story-based analysis of evidence," in *International Conference on Artificial Intelligence* and *Law*, pp. 1-10, 2007.
[71] M. A. Finlayson and P. H. Winston, "Narrative is a key cognitive competency," in *Annual Meeting Biologically Inspired Cognitive Architectures (BICA)*, pp. 110, 2011.
[72] S. Schaal, "The new robotics—towards human-centered machines," *HFSP journal*, vol. 1, no. 2, pp. 115-126, 2007.
[73] T. W. Fong, I. Nourbakhsh, and K. Dautenhahn, "A survey of socially interactive robots," *Robotics* and *Autonomous Systems*, vol. 42, no. 3-4, pp. 143-166, 2002.
[74] S. Kiesler and P. Hinds, "Introduction to this special issue on human-robot interaction," *Human-Computer Interaction*, vol. 19, nos. 1-2, pp. 1-8, 2004.
[75] H. A. Yanco and J. L. Drury, "Classifying human-robot interaction: an updated taxonomy," in *IEEE International Conference on Systems, Man & Cybernetics (SMC)*, pp. 2841-2846, 2004.
[76] A. Bauer, D. Wollherr, and M. Buss, "Human-robot collaboration: A survey," *International Journal of Humanoid Robotics*, vol. 5, no. 1, pp. 47-66, 2008.
[77] M. Asada, K. Hosoda, Y. Kuniyoshi, H. Ishiguro, T. Inui, Y. Yoshikawa, M. Ogino, and C. Yoshida, "Cognitive developmental robotics: A survey," *IEEE Transactions on Autonomous Mental Development*, vol. 1, no. 1, pp. 12-34, 2009.
[78] O. C. Schrempf, U. D. Hanebeck, A. J. Schmid, and H. Worn, "A novel approach to proactive human-robot cooperation," in *IEEE International Workshop on Robot* and *Human Interactive Communication*, pp. 555-560, 2005.
[79] G. Hoffman and C. Breazeal, "Effects of anticipatory action on human-robot teamwork efficiency, fluency, and perception of team," in *Proceedings of the ACM/IEEE international conference on Human-robot interaction*, pp. 1-8, 2007.
[80] P. Tuffield and H. Elias, "The shadow robot mimics human actions," *Industrial Robot: An International Journal*, vol. 30, no. 1, pp. 56-60, 2003.
[81] A. Stenzel, E. Chinellato, M. A. T. Bou, A. P. del Pobil, M. Lappe, and R. Liepelt, "When humanoid robots become human-like interaction partners: corepresentation of robotic actions," *Journal of Experimental Psychology: Human Perception* and *Performance*, vol. 38, no. 5, pp. 1073, 2012.
[82] A. De Santis, B. Siciliano, A. De Luca, and A. Bicchi, "An atlas of physical human–robot interaction," *Mechanism* and *Machine Theory*, vol. 43, no. 3, pp. 253-270, 2008.
[83] D. Kulic and E. A. Croft, "Safe planning for human-robot interaction," *Journal of Robotic Systems*, vol. 22, no. 7, pp. 383-396, 2005.
[84] N. Mitsunaga, C. Smith, T. Kanda, H. Ishiguro, and N. Hagita, "Adapting robot behavior for human--robot interaction," *IEEE Transactions on Robotics*, vol. 24, no. 4, pp. 911-916, 2008.
[85] S. Schaal, "Is imitation learning the route to humanoid robots?" *Trends in Cognitive Science*, vol. 3, no. 6, pp. 233–242, 1999.
[86] M. Mataric, "Getting humanoids to move and imitate," *IEEE Intelligent Systems*, vol. 15, no. 4, pp. 18-24, 2000.
[87] C. D. Frith and D. M. Wolpert, "The neuroscience of social interaction: Decoding, imitating, and influencing the actions of others," *Journal of Consciousness Studies*, vol. 119, no. 4, pp. 664-668, 2004.
[88] R. Stiefelhagen, C. Fugen, R. Gieselmann, H. Holzapfel, K. Nickel, and A. Waibel, "Natural human-robot interaction using speech, head pose and gestures," in *IEEE/RSJ International Conference on Intelligent Robots* and *Systems (IROS)*, 2004.
[89] S. Jerritta, M. Murugappan, R. Nagarajan, and K. Wan, "Physiological signals based human emotion recognition: a review," in *IEEE International Colloquium on Signal Processing* and *its Applications (CSPA)*, 2011.
[90] E. Todorov, "Stochastic optimal control and estimation methods adapted to the noise characteristics of the sensorimotor system," *Neural Computation*, vol. 17, no. 5, pp. 1084–1108, 2005.
[91] R. Liu, X. Zhang, J. Webb, and S. Li, "Context-specific intention awareness through web query in robotic caregiving," in *IEEE International Conference on Robotics* and *Automation (ICRA)*, pp. 1962-1967, 2015.
[92] R. Liu, X. Zhang, and S. Li, "Use context to understand user's implicit intentions in activities of daily living," in *IEEE International Conference on Mechatronics* and *Automation (ICMA)*, pp. 1214-1219, 2014.
[93] R. Liu and X. Zhang, "Fuzzy-context-specific intention inference for robotic caregiving," *International Journal of Advanced Robotic Systems*, vol. 99, no. 99, pp. 99, 2016.
[94] R. Liu and X. Zhang, "Context-specific grounding of web natural descriptions to human-centered situations," *Knowledge-Based Systems*, vol. 111, pp. 1-16, 2016.
[95] A. Burce, I. Nourbakhsh, and R. Simmons, "The role of expressiveness and attention in human-robot interaction," in *IEEE International Conference on Robotics* and *Automation (ICRA)*, vol. 4, pp. 4138-4142, 2002.
[96] M. Staudte and M. W. Crocker, "Investigating joint attention mechanisms through spoken human–robot interaction," *Cognition*, vol. 120, no. 2, pp. 268-291, 2011.
[97] M. L. Walters, K. Dautenhahn, R. Te Boekhorst, K. L. Koay, C. Kaouri, S. Woods, C. Nehaniv, D. Lee, and I. Werry, "The influence of subjects' personality traits on personal spatial zones in a human-robot interaction experiment," in *IEEE International Workshop on Robot* and *Human Interactive Communication*, pp. 347-352, 2005.
[98] L. Takayama and C. Pantofaru, "Influences on proxemic behaviors in human-robot interaction," in *IEEE/RSJ International Conference on Intelligent Robots* and *Systems (IROS)*, pp. 5495-5502, 2009.
[99] R. C. Arkin, M. Fujita, T. Takagi, and R. Hasegawa, "An ethological and emotional basis for human–robot interaction," *Robotics* and *Autonomous Systems*, vol. 42, no. 3, pp. 191-201, 2003.
[100] S. Coradeschi, H. Ishiguro, M. Asada, S. C. Shapiro, M. Thielscher, C. Breazeal, M. J. Mataric, and H. Ishida, "Human-inspired robots," *IEEE Intelligent Systems*, vol. 21, no. 4, pp. 74–85, 2006.
[101] G. Tonietti, R. Schiavi, and A. Bicchi, "Design and control of a variable stiffness actuator for safe and fast physical human/robot interaction," in *IEEE International Conference on Robotics* and *Automation (ICRA)*, pp. 526-531, 2015.
[102] B. Selman. (2014, August 1). NRI: Collaborative Research: Jointly Learning Language and Affordances. Available: https://www.nsf.gov/awardsearch/showAward?AWD_ID=1426744&HistoricalAwards=false
[103] R. Mooney. (2016, August 1). NRI: Robots that Learn to Communicate through Natural Human Dialog. Available: https://www.nsf.gov/awardsearch/showAward?AWD_ID=1637736&HistoricalAwards=false
[104] N. Roy. (2014, August 18). NRI: Collaborative Research: Modeling and Verification of Language-based Interaction. Available: https://www.nsf.gov/awardsearch/showAward?AWD_ID=1427547&HistoricalAwards=false
[105] University of Washington. (2017, January 5). Robotics and State Estimation Lab. Available: http://rse-lab.cs.washington.edu/projects/language-grounding/
[106] Lund University. (2017, January 5). Robotics and State Estimation Lab. Available: http://rss.cs.lth.se/
[107] J. Hollerbach, "The international journal of robotics research," SAGE. [Online]. Available: http://journals.sagepub.com/home/ijr. Accessed: Jan. 27, 2017.
[108] F. Park, "IEEE Xplore: IEEE transactions on robotics," in IEEE transactions on robotics. [Online]. Available: http://ieeexplore.ieee.org/xpl/RecentIssue.jsp?punumber=8860. Accessed: Jan. 27, 2017.
[109] R. Dechter, Artificial intelligence. [Online]. Available: https://www.journals.elsevier.com/artificial-intelligence/. Accessed: Jan. 27, 2017.

[110]  H. Fujita and J. Lu, Knowledge-based systems. [Online]. Available: https://www.journals.elsevier.com/knowledge-based-systems. Accessed: Jan. 27, 2017.
[111]  "ICRA2015," in ICRA2015. [Online]. Available: http://ieeexplore.ieee.org/xpl/mostRecentIssue.jsp?punumber=7128761. Accessed: Jan. 27, 2017.
[112]  "IROS2015," in IROS2015. [Online]. Available: https://www.ieee.org/conferences_events/conferences/conferencedetails/index.html?Conf_ID=33365. Accessed: Jan. 27, 2017.
[113]  "AAAI15: Twenty-Ninth conference on artificial intelligence," in AAAI15, 1995. [Online]. Available: http://www.aaai.org/Conferences/AAAI/aaai15.php. Accessed: Jan. 27, 2017.
[114]  B. D. Argall, S. Chernova, M. Veloso, and B. Browning, "A survey of robot learning from demonstration," *Robotics* and *autonomous systems*, vol. 57, no. 5, pp. 469-483, 2009.
[115]  S. S. Rautaray and A. Agrawal, "Vision based hand gesture recognition for human computer interaction: a survey," *Artificial Intelligence Review*, vol. 43, no. 1, pp. 1-54, 2015.
[116]  C. L. Bethel, K. Salomon, R. R. Murphy, and J. L. Burke, "Survey of psychophysiology measurements applied to human-robot interaction," in *IEEE International Symposium on Robot* and *Human Interactive Communication*, pp. 732-737, 2007.
[117]  B. D. Argall and A. G. Billard, "A survey of tactile human–robot interactions," *Robotics* and *Autonomous Systems*, vol. 58, no. 10, pp. 1159-1176, 2010.
[118]  G. Antoniol, R. Cattoni, M. Cettolo, and M. Federico, "Robust speech understanding for robot telecontrol," in *International Conference on Advanced robotics*, pp. 205-209,1993.
[119]  S. Chen, Z. Kazi, M. Beitler, M. Salganicoff, D. Chester, and R. Foulds, "Gesture-speech based HMI for a rehabilitation robot," in *Proceedings of Bringing Together Education, Science* and *Technology*, pp.29-36, 1996.
[120]  R. Bischoff and V. Graefe, "Integrating vision, touch and natural language in the control of a situation-oriented behavior-based humanoid robot," in *IEEE International Conference on Systems, Man,* and *Cybernetics (SMC)*, vol.2, pp. 999-1004, 1999.
[121]  B. Landau and R. Jackendoff, "Whence and whither in spatial language and spatial cognition?" *Behavioral* and *brain sciences*, vol. 16, no. 2, pp. 255-265, 1993.
[122]  M. Ferre, J. Macias-Guarasa, R. Aracil, and A. Barrientos, "Voice command generation for teleoperated robot systems," in *IEEE International Workshop on Robot* and *Human Communication*, pp. 679-685, 1998.
[123]  J. Savage, E. Hernandez, G. Vazquez, A. Hernandez, and A. L. Ronzhin, "Control of a mobile robot using spoken commands," in *Conference Speech* and *Computer*, pp. 333-338, 2004.
[124]  C. Jayawardena, K. Watanabe, and K. Izumi, "Posture control of robot manipulators with fuzzy voice commands using a fuzzy coach–player system," *Advanced Robotics*, vol. 21, nos. 3-4, pp. 293-328, 2007.
[125]  A. Nordmann, S. Wrede, and J. Steil, "Modeling of movement control architectures based on motion primitives using domain-specific language," in *IEEE International Conference on Robotics* and *Automation (ICRA)*, pp. 5032-5039, 2015.
[126]  M. Beetz, F. Balint-Benczedi, N. Blodow, D. Nyga, T. Wiedemeyer, and Z. C. Marton, "ROBOSHERLOCK: Unstructured information processing for robot perception," in *IEEE International Conference on Robotics* and *Automation (ICRA)*, pp. 1549-1556, 2015.
[127]  J. Allen, Q. Duong, and C. Thompson, "Natural language service for controlling robots and other agents," in *IEEE Integration of Knowledge Intensive Multi-Agent Systems (KIMAS-05)*, pp. 592-595, 2005.
[128]  B. House, J. Malkin, and J. Bilmes, "The VoiceBot: A Voice Controlled Robot Arm," *In Proceedings of the SIGCHI Conference on Human Factors in Computing Systems*, pp. 183-192, 2009.
[129]  M. Stenmark and P. Nugues, "Natural Language Programming of Industrial Robots," International Symposium on Robotics, pp. 1-5, 2013.
[130]  D. Jain, L. Mosenlechner, and M. Beetz, "Equipping robot control programs with first-order probabilistic reasoning capabilities," in *IEEE International Conference on Robotics* and *Automation (ICRA)*, pp. 3626-3631, 2009.
[131]  G. E. Fainekos, H. Kress-Gazit, and G. J. Pappas, "Temporal logic motion planning for mobile robots," in *IEEE International Conference on Robotics* and *Automation (ICRA)*, pp. 2020-2025, 2005.
[132]  T. Oates, Z. Eyler-Walker, and P. R. Cohen, "Using syntax to learn semantics: An experiment in language acquisition with a mobile robot," *Walker*, pp. 1-6, 1999.
[133]  M. Stenmark, J. Malec, K. Nilsson, and A. Robertsson, "On distributed knowledge bases for robotized small-batch assembly," *IEEE Transactions on Automation Science* and *Engineering*, vol. 12, no. 2, pp. 1-10, 2015.
[134]  J. M. G. Romano, E. F. Camacho, J. G. Ortega, and M. T. Bonilla, "A generic natural language interface for task planning application to a mobile robot," *Control Engineering Practice*, vol. 8, no. 10, pp. 1119-1133, 2000.
[135]  C. Finucane, G. Jing, and H. Kress-Gazit, "LTLMoP: Experimenting with language, temporal logic and robot control," in *IEEE/RSJ International Conference on Intelligent Robots* and *Systems (IROS)*, pp. 1988-1993, 2010.
[136]  M. Bollini, S. Tellex, T. Thompson, N. Roy, and D. Rus, "Interpreting and executing recipes with a cooking robot," in *13th International Symposium on Experimental Robotics*, pp. 481-495, 2013.
[137]  H. Cuayahuitl, "Robot learning from verbal interaction: A brief survey," in *International Symposium on New Frontiers in Human-Robot Interaction*, 2015.
[138]  Y. Chen and D. H. Ballard, "On the integration of grounding language and learning objects," in *National Conference on Artificial Intelligence*, vol. 4, pp. 2, 2004.
[139]  M. N. Nicolescu and M. J. Matari, "Task learning through imitation and human-robot interaction," in *Models* and *Mechanisms of Imitation* and *Social Learning in Robots, Humans* and *Animals: Behavioural, Social* and *Communicative Dimension*, pp. 407-424, 2005.
[140]  D. Roy, "Learning visually grounded words and syntax of natural spoken language," *Evolution of Communication*, vol. 4, no. 1, pp. 33-56, 2002.
[141]  S. Lauria, G. Bugmann, T. Kyriacou, and J. Bos, "Personal Robot Training via Natural-Language Instructions," IEEE Intelligent Systems, vol. 16, no.5, pp. 38-45, 2001.
[142]  M. N. Nicolescu and M. J. Mataric, "Natural methods for robot task learning: instructive demonstrations, generalization and practice," in *Proceedings of the second international joint conference on Autonomous agents* and *multiagent systems*, pp. 241-248, 2003.
[143]  G. M. Kruijff, J. D. Kelleher, G. Berginc, and A. Leonardis, "Structural descriptions in human-assisted robot visual learning," in *ACM SIGCHI/SIGART conference on Human-robot interaction (HRI)*, pp. 343-344, 2006.
[144]  K. Sugiura and N. Iwahashi, "Learning object-manipulation verbs for human-robot communication," in *Proceedings of the 2007 workshop on Multimodal interfaces in* semantic *interaction (WMISI)*, pp. 32-38, 2007.
[145]  P. Kordjamshidi, J. Hois, M. V. Otterlo, and M. Moens, "Learning to interpret spatial natural language in terms of qualitative spatial relations," *Representing space in cognition: interrelations of behavior, language,* and *formal models*, 2012.
[146]  N. Iwahashi, "Robots that learn language: A developmental approach to situated human-robot conversations," in *Human Robot Interaction*, Nilanjan Sarkar (Ed.), InTech, 2007.
[147]  K. F. Uyanik, Y. Caliskan, A. K. Bozcuoglu, O. Yuruten, S. Kalkan, and E. Sahin, "Learning social affordances and using them for planning," in *35th Annual Meeting of the Cognitive Science Society (CogSci)*, 2013.
[148]  A. Holroyd and C. Rich, "Using the behavior markup language for human-robot interaction," in *Proceedings of the seventh annual ACM/IEEE international conference on Human-Robot Interaction*, pp. 147-148, 2012.
[149]  L. Montesano, M. Lopes, A. Bernardino, and J. Santos-Victor, "Modeling affordances using Bayesian networks," in *IEEE/RSJ International Conference on Intelligent Robots* and *Systems (IROS)*, pp. 4102-4107, 2007.
[150]  M. Salem, S. Kopp, I. Wachsmuth, and F. Joublin, "Towards an integrated model of speech and gesture production for multi-modal robot behavior," in *IEEE International Symposium on Robot* and *Human Interactive Communication (RO-MAN)*, pp. 614-619, 2010.
[151]  M. Forbes, M. Chung, M. Cakmak, and L. Zettlemoyer, "Grounding antonym adjective pairs through interaction," HRI, 2014.
[152]  E. Krause, M. Zillich, T. Williams, and M. Scheutz, "Learning to recognize novel objects in one shot through human-robot interactions in natural language dialogues," in *Proceedings of the Twenty-Eighth AAAI Conference on Artificial Intelligence*, pp. 2796-2802, 2014.
[153]  H. Dindo and D. Zambuto, "A probabilistic approach to learning a visually grounded language model through human-robot interaction," in *IEEE/RSJ International Conference on Intelligent Robots* and *Systems (IROS)*, pp. 790-796, 2010.

[154] T. Williams, G. Briggs, B. Oosterveld, and M. Scheutz, "Going beyond literal command-based instructions: Extending robotic natural language interaction capabilities," in *AAAI Conference on Artificial Intelligence*, pp. 1387-1393, 2015.
[155] A. Bannat, J. Blume, J. T. Geiger, T. Rehrl, and F. Wallhoff, C. Mayer, B. Radig, S. Sosnowski, and K. Kuhnlenz, "A multimodal human-robot-dialog applying emotional feedbacks," in *Proceedings of Second International Conference on Social Robotics (ICSR)*, 2010.
[156] A. L. Thomaz and C. Breazeal, "Teachable robots: Understanding human teaching behavior to build more effective robot learners," *Artificial Intelligence*, vol. 172, nos. 6-7, pp. 716-737, 2008.
[157] J. R. Medina, M. Lawitzky, A. Mortl, and D. Lee, "An experience-driven robotic assistant acquiring human knowledge to improve haptic cooperation," in *IEEE/RSJ International Conference on Intelligent Robots and Systems (IROS)*, 2011.
[158] K. Sugiura, N. Iwahashi, H. Kawai, and S. Nakamura, "Situated spoken dialogue with robots using active learning," *Advanced Robotics*, vol. 25, no. 17, pp. 2207-2232, 2011.
[159] D. K. Misra, J. Sung, K. Lee, and A. Saxena, "Tell me Dave: Context-sensitive grounding of natural language to manipulation instructions," in *Robotics: Science and Systems*, vol.35, no. 1, pp. 281-300, 2014.
[160] E. Bicho, L. Louro, W. Erlhagen, "Integrating verbal and nonverbal communication in a dynamic neural field architecture for human–robot interaction," Frontiers in neurorobotics, vol. 4, no. 5, 2010.
[161] A. Broad, J. Arkin, N. Ratliff, T. Howard, and B. Argall, "Towards Real-Time Natural Language Corrections for Assistive Robots," RSS Workshop on Model Learning for Human-Robot Communication, 2016.
[162] P. F. Dominey, A. Mallet, and E. Yoshida, "Progress in programming the hrp-2 humanoid using spoken language," in *IEEE International Conference on Robotics and Automation (ICRA)*, pp. 2169-2174, 2007.
[163] S. Profanter, A. Perzylo, N. Somani, M. Rickert, and A. Knoll, "Analysis and semantic modeling of modality preferences in industrial human-robot interaction," in *IEEE/RSJ International Conference on Intelligent Robots and Systems (IROS)*, pp. 1812-1818, 2015.
[164] D. Lu, F. Wu, and X. Chen, "Understanding User Instructions by Utilizing Open Knowledge for Service Robots," arXiv preprint arXiv:1606.02877, 2016.
[165] B. Thomas, and O. Jenkins, "RoboFrameNet: Verb-Centric Semantics for Actions in Robot Middleware," IEEE International Conference on Robotics and Automation, pp. 4750-4755, 2012.
[166] B. Burger, I. Ferrane, F. Lerasle, and G. Infantes, "Two-handed gesture recognition and fusion with speech to command a robot," *Autonomous Robots*, vol. 32, no. 2, pp. 129-147, 2012.
[167] T. Fong, I. Nourbakhsh, C Kunz, L. Fluckiger and J. Schreiner, "The Peer-to-Peer Human-Robot Interaction Project," Space 2005, pp. 6750, 2005.
[168] R. Deits, S. Tellex, P. Thaker, D. Simeonov, T. Kolloar, and N. Roy, "Clarifying commands with information-theoretic human-robot dialog," *Journal of Human-Robot Interaction*, vol. 1, no. 1, pp. 78-95, 2012.
[169] P. Rybski, J. Stolarz, K. Yoon and M. Veloso, "Using dialog and human observations to dictate tasks to a learning robot assistant," Intelligent Service Robotics, vol. 1, no. 2, pp. 159-167, 2008.
[170] R. Bischoff and V. Graefe, "Dependable multimodal communication and interaction with robotic assistants," in *IEEE International Workshop on Robot and Human Interactive Communication (RO-MAN)*, pp. 300-305, 2002.
[171] A. Clodic, R. Alami, V. Montreuil, and S. Li, "A study of interaction between dialog and decision for human-robot collaborative task achievement," in *International Symposium on Robot and Human interactive Communication (RO-MAN)*, pp. 913-918, 2007.
[172] S. S. Ghidary, Y. Nakata, H. Saito, M. Hattori, and T. Takamori, "Multi-Modal human robot interaction for map generation," in *IEEE/RSJ International Conference on Intelligent Robots and Systems (IROS)*, vol. 4, pp. 2246-2251, 2001.
[173] T. Kollar, S. Tellex, D. Roy, and N. Roy, "Grounding verbs of motion in natural language commands to robots," in *12th International Symposium on Experimental Robotics*, pp. 31-47, 2014.
[174] J. Bos, "Applying automated deduction to natural language understanding," *Journal of Applied Logic*, vol. 7, no. 1, pp. 100-112, 2009.
[175] M. Mason and M. Lopes, "Robot self-initiative and personalization by learning through repeated interactions," In Proceedings of the 6th international conference on Human-robot interaction, pp. 433-440, 2011.
[176] M. Strait, P. Briggs, and M. Scheutz, "Gender, more so than Age, Modulates Positive Perceptions of Language-Based Human-Robot Interactions," International Symposium on New Frontiers in Human-Robot Interaction, 2015.
[177] J. F. Gorostiza and M. A. Salichs, "Natural programming of a social robot by dialogs," in *AAAI Fall Symposium*, 2010.
[178] B. Mutlu, J. Forlizzi and J. Hodgins, "A Storytelling Robot: Modeling and Evaluation of Human-like Gaze Behavior," IEEE-RAS International conference on humanoid robotics, pp. 518-523, 2006.
[179] W. Wang, G. Athanasopoulos, S. Yilmazyildiz, G. Patsis, V. Enescu, H. Sahli, W. Verhelst, A. Hiolle, M. Lewis, and L. Canamero, "Natural emotion elicitation for emotion modeling in child-robot interactions," in *4th Workshop on Child Computer Interaction (WOCCI)*, pp. 51-56, 2014.
[180] C. Breazeal and L. Aryananda, "Recognition of affective communicative intent in robot-directed speech," *Autonomous Robots*, vol. 12, no. 1, pp. 83-104, 2002.
[181] A. Lockerd and C. Breazeal, "Tutelage and socially guided robot learning," in *IEEE/RSJ International Conference on Intelligent Robots and Systems (IROS)*, vol. 4, pp. 3475-3480, 2004.
[182] C. Breazeal, "Toward sociable robots," *Robotics and Autonomous Systems*, vol. 42, no. 3, pp. 167-175, 2003.
[183] K. Severinson-Eklundh, A. Green, and H. Huttenrauch, "Social and collaborative aspects of interaction with a service robot," *Robotics and Autonomous Systems*, vol. 42, nos. 3-4, pp. 223-234, 2003.
[184] A. Austermann, N. Esau, L. Kleinjohann, and B. Kleinjohann, "Fuzzy emotion recognition in natural speech dialogue," in *International Workshop on Robot and Human interactive Communication (RO-MAN)*, pp. 317-332, 2005.
[185] M. Coeckelbergh, "You, robot: on the linguistic construction of artificial others," *AI & Society*, vol. 26, no. 1, pp. 61-69, 2011.
[186] R. Read and T. Belpaeme, "How to use non-linguistic utterances to convey emotion in child-robot interaction," in *7th ACM/IEEE International Conference on Human-Robot Interaction (HRI)*, pp. 219-220, 2012.
[187] I. Kruijff-Korbayova, I. Baroni, and M. Nalin, "Children's Turn-Taking Behavior Adaptation in Multi-Session Interactions with a Humanoid Robot," In IEEE workshop on timing in human-robot interaction, 2014.
[188] S. Sabanovic, M. P. Michalowski, and R. Simmons, "Robots in the wild: Observing human-robot social interaction outside the lab," in *9th IEEE International Workshop on Advanced Motion Control*, pp. 596-601, 2006.
[189] H. G. Okuno, K. Nakadai, and H. Kitano, "Social interaction of humanoid robot based on audio-visual tracking," in *International Conference on Industrial and Engineering Applications of Artificial Intelligence and Expert Systems*, pp. 725-735, 2002.
[190] A. Chella, R. E. Barone, G. Pilato, and R. Sorbello, "An emotional storyteller robot," in *AAAI Spring Symposium on Emotion, Personality, and Social Behavior*, pp. 17-22, 2008
[191] T. Kollar, S. Tellex, M. R. Walter, A. Huang, A. Bachrach, S. Hemachandra, E. Brunskill, A. Banerjee, D. Roy, S. Teller, and N. Roy, "Generalized grounding graphs: A probabilistic framework for understanding grounded language," *Journal of Artificial Intelligence Research*, pp. 1-35, 2013.
[192] M. Forbes, R. P. N. Rao, L. Zettlemoyer, and M. Cakmak, "Robot programming by demonstration with situated spatial language understanding," in *IEEE International Conference on Robotics and Automation (ICRA)*, pp. 2014-2020, 2015.
[193] G. Salvi, L. Montesano, A. Bernardino, and J. Santos-Victor, "Language bootstrapping: Learning word meanings from perception–action association," *IEEE Transactions on Systems, Man, and Cybernetics, Part B (Cybernetics)*, vol. 42, no. 3, pp. 660-671, 2011.
[194] C. M. Huang and B. Mutlu, "Anticipatory robot control for efficient human-robot collaboration," in *11th ACM/IEEE International Conference on Human-Robot Interaction (HRI)*, pp. 83-90, 2016.
[195] M. Johnson-Roberson, J. Bohg, G. Skantze, J. Gustafson, R. Carlson, B. Rasolzadeh, and D. Kragic, "Enhanced visual scene understanding through human-robot dialog," in *IEEE/RSJ International Conference on Intelligent Robots and Systems (IROS)*, pp. 3342-3348, 2011.
[196] F. Doshi and N. Roy, "Spoken language interaction with model uncertainty: an adaptive human–robot interaction system," *Connection Science*, vol. 20, no. 4, pp. 299-318, 2008.

[197] S. Lauria, G. Bugmann, T. Kyriacou, and E. Klein, "Mobile robot programming using natural language," *Robotics* and *Autonomous Systems*, vol. 38, no. 3-4, pp. 171-181, 2002.
[198] S. Hemachandra, M. R. Walter, S. Tellex, and S. Teller, "Learning spatial-semantic representations from natural language descriptions and scene classifications," in *IEEE International Conference on Robotics* and *Automation (ICRA)*, pp. 2623-2630, 2015.
[199] M. Hanheide, M. Gobelbecker, G. S. Horn, A. Pronobis, K. Sjoo, A. Aydemir, P. Jensfelt, C. Gretton, R. Dearden, M. Janicek, H. Zender, G. Kruijff, N. Hawes, and J. L. Wyatt, "Robot task planning and explanation in open and uncertain worlds," *Artificial Intelligence*, 2015.
[200] X. Chen, J. Xie, J. Ji, and Z. Sui, "Toward open knowledge enabling for Human-Robot interaction," *Journal of Human-Robot Interaction*, vol. 1, no. 2, pp. 100-117, 2012.
[201] J. Weston, F. Ratle, and R. Collobert, "Deep learning via semi-supervised embedding," in *Proceedings of the 25th international conference on Machine learning (ICML)*, pp. 639-655, 2008.
[202] I. Sutskever, O. Vinyals, and Q. V. Le, "Sequence to sequence learning with neural networks," *Advances in neural information processing systems*, vol. 4, pp. 3104-3112, 2014.
[203] D. Bahdanau, K. Cho, and Y. Bengio, "Neural machine translation by jointly learning to align and translate," in *International Conference on Learning Representation (ICLR)*, 2015.
[204] X. Glorot, A. Bordes, and Y. Bengio, "Domain adaptation for large-scale sentiment classification: A deep learning approach," in *International Conference on Machine learning (ICML)*, pp. 513-520, 2011.
[205] M. Samadi, T. Kollar, and M. Veloso, "Using the web to interactively learn to find objects," in *Proceedings of the Twenty-Sixth AAAI Conference on Artificial Intelligence*, 2012.
[206] G. Gordon, S. Spaulding, J. K. Westlund, J. L. Jin, L. Plummer, M. Martinez, M. Das, and C. Breazeal, "Affective personalization of a social robot tutor for children's second language skills," in *Proceedings of the Thirtieth AAAI Conference on Artificial Intelligence*, pp. 3951-3957, 2016.
[207] J. Saunders, D. S. Syrdal, K. L. Koay, N. Burke, and K. Dautenhahn, "Teach me -- show me" -- end-user personalization of a smart home and companion robot," *IEEE Transactions on Human-Machine Systems*, vol. 46, no. 1, pp. 27-40, 2016.
[208] M. Slamani, A. Nubiola, and I. Bonev, "Assessment of the positioning performance of an industrial robot," *Industrial Robot*, vol. 39, no. 1, pp. 57-68, 2012.
[209] A. Antunes, L. Jamone, G. Saponaro, A. Bernardino, and R. Ventura, "From human instructions to robot actions: Formulation of goals, affordances and probabilistic planning," in *IEEE International Conference on Robotics* and *Automation (ICRA)*, pp. 5449-5454, 2016.
[210] G. Gordon and C. Breazeal, "Bayesian active learning-based robot tutor for children's word-reading skills," in *Proceedings of the Twenty-Ninth AAAI Conference on Artificial Intelligence*, pp. 1343-1349, 2015.
[211] Q. Zeng, S. Sun, H. Duan, C. Liu, and H. Wang, "Cross-organizational collaborative workflow mining from a multi-source log," *Decision Support Systems* vol. 54, no. 3, pp. 1280-1301, 2013.
[212] R. Liu, X. Zhang, and H. Zhang, "Web-video-mining-supported workflow modeling for laparoscopic surgeries," Artificial Intelligence in Medicine, vol. 74, pp. 9-20, 2016.
[213] R. Liu and X. Zhang, "Generating machine-executable plans from end-user's natural-language-instructions," arXiv preprint arXiv:1611.06468, 2016.
[214] N. Legany, G. Toldi, N. Megyes, C. Orban, L. Kovacs, and A. Balog, "ACT-R/E: An embodied cognitive architecture for human-robot interaction," *Journal of Human-Robot Interaction*, vol. 2, no. 1, pp. 30-55, 2013.
[215] H. Koppula and A. Saxena, "Anticipating human activities using object affordances for reactive robotic response," in *Robotics: Science* and *Systems*, vol. 38, no. 1, pp. 14-29, 2013.
[216] M. L. Littman, "Reinforcement learning improves behaviour from evaluative feedback," *Nature*, vol. 521, no. 7553, pp. 445-451, 2015.
[217] P. Raccuglia, K. C. Elbert, P. D. F. Adler, C. Falk, M. B. Wenny, A. Mollo, M. Zeller, S. A. Friedler, J. Schrier, and A. J. Norquist, "Machine-learning-assisted materials discovery using failed experiments," Nature, vol. 533, no. 7601, pp.73-76, 2016.

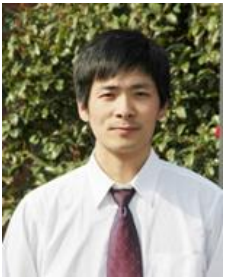
**Rui Liu** received the M.S. degree in active vibration control from Shanghai Jiao Tong University, Shanghai, China in 2013. Since January, 2014, he has been a Ph.D. student in Mechanical Engineering at Colorado School of Mines, Golden, Colorado, USA.

From June to December 2013, he was a Research Engineer at the Chinese Academy of Science, Chengdu, China. His current research interests include robotics, Knowledge-based Systems, NLP and machine learning.

All his papers could be found at:
Google Scholar: https://scholar.google.com/citations?user=X5X0IyQAAAAJ&hl=en&authuser=1
Research Gate: https://www.researchgate.net/profile/Rui_Liu99/contributions

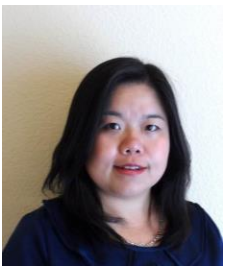
**Xiaoli Zhang** received the B.S. degree in Mechanical Engineering from Xi'an Jiaotong University, Xi'an, ShanXi, China in 2003, the M.S. degree in Mechatronics Engineering from Xi'an Jiaotong University in 2006, the Ph.D. degree in Biomedical Engineering from the University of Nebraska Lincoln, Lincoln, USA, in 2009.

Since 2013, she has been an Assistant Professor in the Mechanical Engineering Department, Colorado School of Mines, Golden, CO. She is the author of more than 50 articles, and 5 inventions. Her research interests include intelligent human-robot interaction, human intention awareness, robotics system design and control, haptics, and their applications in healthcare fields.